\documentclass[runningheads]{llncs}

\usepackage[T1]{fontenc}
\usepackage[utf8]{inputenc}

\usepackage{graphicx}

\usepackage{quantikz}

\usepackage{booktabs}
\usepackage{tabularx}
\usepackage{caption}

\usepackage{amsmath, amssymb}

\usepackage{microtype}

\usepackage{tikz}
\usetikzlibrary{positioning,calc}
\usepackage{placeins}

\usepackage{subcaption}

\usepackage{braket}

\usepackage{hyperref}

\begin{document}

\title{Comparing Classical and Quantum Variational Classifiers on the XOR Problem}
\titlerunning{Classical and Quantum Variational Classification of XOR}

\author{Miras Seilkhan\inst{1} \and Adilbek Taizhanov\inst{2}}
\authorrunning{M. Seilkhan and A. Taizhanov}

\institute{
\email{seilkhan.miras6117@gmail.com}\\
\and
\email{adilbek300108@gmail.com}
}

\maketitle
\begin{abstract}

Quantum machine learning applies principles of quantum mechanics, such as superposition and entanglement, to data processing and optimization. Variational quantum models operate on qubits in high-dimensional Hilbert spaces and offer an alternative route to model expressivity.

We compare classical models and a variational quantum classifier on the XOR problem. Logistic regression (LR), a one-hidden-layer multilayer perceptron (MLP), and a two-qubit VQC with circuit depths $L\in\{1,2\}$ are evaluated on synthetic XOR datasets with varying Gaussian noise and sample sizes using accuracy and binary cross-entropy.

Performance is driven by model expressivity: LR and VQC($L=1$) fail to represent XOR reliably, whereas MLP and VQC($L=2$) achieve perfect test accuracy on representative settings (e.g., $\sigma=0.10$, $n=100$ per cluster). Robustness analyses over noise levels, dataset sizes, and random seeds confirm that circuit depth is decisive for the VQC in this task. Despite identical maximal accuracy, MLP attains lower binary cross-entropy and substantially lower training time. Hardware inference preserves the global XOR structure but introduces structured deviations in the decision function (mean absolute deviation $\approx 0.118$).

Overall, deeper VQCs match classical neural networks in accuracy on low-dimensional XOR benchmarks but show no clear empirical advantage in robustness or efficiency within the examined configurations.

\end{abstract}

{
\setlength{\parskip}{0.4em}

\section{Introduction}
Classical machine learning is a set of statistical algorithms and methods aimed at analyzing data and identifying patterns based on experience. This approach has been successfully applied to many practical problems, but its predictive capabilities are significantly limited by the model architecture and the data representation methods used. In particular, linear classifiers and single-layer models are unable to solve linearly inseparable problems without explicitly introducing nonlinear transformations of the feature space.
\newpage
Quantum machine learning offers an alternative computational paradigm based on principles of quantum mechanics, such as superposition and entanglement. Encoding classical data into states of quantum systems allows information to be represented in multidimensional Hilbert spaces, opening the possibility of implementing nonlinear transformations using compact, parameterized quantum circuits. Variational quantum classifiers represent a hybrid approach in which quantum circuit parameters are iteratively optimized using classical methods, making such models applicable to modern intermediate-scale quantum devices.

To assess the capabilities of quantum algorithms, it is essential to consider problems with a well-defined nonlinear structure. The XOR (exclusive OR) problem is a minimal example of a linearly inseparable binary classification task and is widely used in machine learning and neural network theory to illustrate the limitations of simple models and the necessity of nonlinear representations.

In this work, we present a comparative analysis of classical machine learning models and a variational quantum classifier using the XOR problem as a benchmark. We examine the impact of noise level, training set size, random parameter initialization, and architectural choices on classification performance. Experiments are conducted both in idealized quantum simulation and on real quantum hardware, allowing us to assess the effects of quantum noise and hardware constraints on training outcomes.
}

{
\setlength{\parskip}{0.4em}

\section{Background / Theory}
\label{sec:background}

This section introduces the classical and quantum models used in our experiments and formalizes why XOR is a minimal example of a \emph{linearly inseparable} classification task. We consider inputs $x=(x_1,x_2)^\top \in \mathbb{R}^2$ and binary labels $y\in\{0,1\}$.

\subsection{Classical machine learning models}
\label{subsec:classical}
\subsubsection{Linear classifier}
\label{subsubsec:linear}

A \emph{linear classifier} is one of the simplest models for binary classification.
Its decision boundary is a hyperplane in the original feature space
(a straight line in $\mathbb{R}^2$).
A typical example of such a model is logistic regression.

Given an input vector $x=(x_1,x_2)^\top \in \mathbb{R}^2$,
the classifier computes a linear score
\begin{equation}
z(x) = w^\top x + b,
\end{equation}

where $w\in\mathbb{R}^2$ is the weight vector and $b\in\mathbb{R}$ is the bias term. Geometrically, the equation $w^\top x + b = 0$ defines a straight line that separates
the feature space into two regions.

To obtain a probabilistic interpretation, the score $z(x)$ is passed through
the logistic sigmoid function
\begin{equation}
p_\theta(y=1 \mid x) = \sigma\!\bigl(z(x)\bigr)
= \frac{1}{1+e^{-z(x)}},
\qquad \theta = (w,b),
\end{equation}

which maps real-valued inputs to the interval $(0,1)$.
Predicted class labels are obtained by thresholding the probability at $0.5$:
\begin{equation}
\hat{y}(x)
= \mathbb{I}\!\left[p_\theta(y=1 \mid x) > 0.5\right]
= \mathbb{I}\!\left[z(x) > 0\right],
\end{equation}

where $\mathbb{I}[\cdot]$ denotes the indicator function, which returns $1$ if the
condition is satisfied and $0$ otherwise.

Since the prediction rule depends only on the sign of the linear function
$w^\top x + b$, a linear classifier can generate only linear decision boundaries.
As a result, such models are fundamentally limited and cannot solve linearly
inseparable problems such as the XOR classification task.

\begin{figure}[h]
\centering
\includegraphics[width=0.55\linewidth]{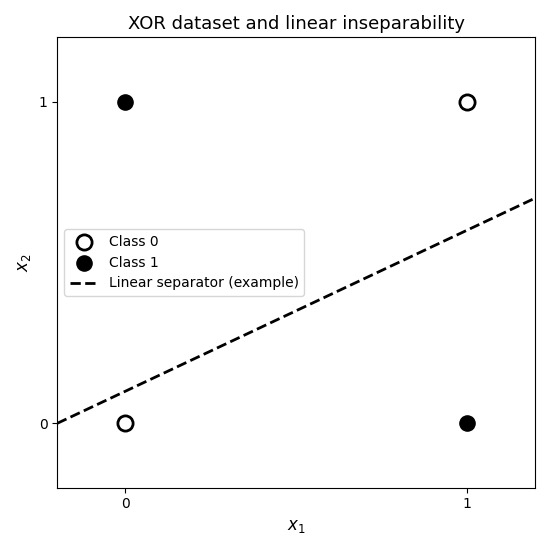}
\caption{Illustration of the XOR classification problem.
The two classes (black and white points) cannot be separated by a single straight line,
demonstrating the linear inseparability of XOR.}
\label{fig:xor_inseparable}
\end{figure}

Formally, the decision boundary of a linear classifier is given by the set
\[
\{x : w^\top x + b = 0\},
\]

which defines a hyperplane in the feature space (a line in $\mathbb{R}^2$).

To learn the model parameters $\theta = (w,b)$ from data, we minimize the empirical cross-entropy loss (also known as log-loss), defined as
\begin{equation}
\mathcal{L}(\theta) = -\frac{1}{N}\sum_{i=1}^N \left[
y_i \log p_\theta(y=1 \mid x_i)
+ (1-y_i)\log\bigl(1-p_\theta(y=1 \mid x_i)\bigr)
\right].
\label{eq:ce_linear}
\end{equation}

\subsubsection{Why a linear model cannot solve XOR}
\label{subsubsec:xor_proof}

Consider the canonical XOR dataset consisting of four input-output pairs:
\[
(0,0)\mapsto 0,\quad (0,1)\mapsto 1,\quad (1,0)\mapsto 1,\quad (1,1)\mapsto 0.
\]
We assume, for the sake of contradiction, that there exists a linear classifier
\[
\hat{y}(x) = \mathbb{I}[w^\top x + b > 0]
\]
that correctly classifies all four points.
This assumption imposes the following constraints on the model parameters:
\begin{align}
(0,0)\to 0 &\Rightarrow b < 0, \label{eq:xor1}\\
(1,0)\to 1 &\Rightarrow w_1 + b > 0 \Rightarrow w_1 > -b, \label{eq:xor2}\\
(0,1)\to 1 &\Rightarrow w_2 + b > 0 \Rightarrow w_2 > -b, \label{eq:xor3}\\
(1,1)\to 0 &\Rightarrow w_1 + w_2 + b < 0 \Rightarrow w_1 + w_2 < -b. \label{eq:xor4}
\end{align}
From \eqref{eq:xor2} and \eqref{eq:xor3}, it follows that:
\[
w_1 + w_2 > -2b.
\]
Since \eqref{eq:xor1} implies $b<0$, we have $-2b > -b$, and therefore
\[
w_1 + w_2 > -b,
\]
which contradicts condition \eqref{eq:xor4}. Hence, no linear classifier can separate the XOR dataset, and the XOR problem is linearly inseparable, as originally observed by Minsky and Papert~\cite{minsky1988}.

\subsubsection{Multilayer Perceptron (MLP)}
\label{subsubsec:mlp}

Unlike linear classifiers, a multilayer perceptron (MLP) can model nonlinear relationships by introducing hidden layers with nonlinear activation functions. In this work, we consider a one-hidden-layer MLP with $h$ hidden units, where $h$ is treated as a hyperparameter and varied in the experiments.

The hidden layer applies a linear transformation to the input vector
$x \in \mathbb{R}^2$, followed by a nonlinear activation:
\begin{equation}
a(x) = \phi\!\left(W_1 x + b_1\right),
\label{eq:mlp_hidden}
\end{equation}
where $W_1 \in \mathbb{R}^{h \times 2}$ is the weight matrix of the hidden layer, $b_1 \in \mathbb{R}^h$ is the corresponding bias vector, and $\phi(\cdot)$ denotes an element-wise nonlinear activation function such as $\tanh$ or ReLU. The vector 
$a(x) \in \mathbb{R}^h$ represents the activations of the hidden units.

The output layer then maps these activations to a scalar score:
\begin{equation}
z(x) = W_2 a(x) + b_2,
\label{eq:mlp_score}
\end{equation}
where $W_2 \in \mathbb{R}^{1 \times h}$ and $b_2 \in \mathbb{R}$ are the weights and bias of the output layer.

Finally, the predicted probability of assigning the input $x$ to class $1$ is obtained by applying the sigmoid function:
\begin{equation}
p_\theta(y=1 \mid x) = \sigma\!\left(z(x)\right),
\label{eq:mlp_prob}
\end{equation}
where $\sigma(\cdot)$ ensures that the output lies in the interval $(0,1)$ and
$\theta = (W_1, b_1, W_2, b_2)$ denotes the full set of trainable model parameters.

By combining linear transformations with nonlinear activations, the MLP can form nonlinear decision boundaries. As a result, even a small MLP can successfully solve the XOR classification problem.
{
\begin{figure}[h]
\centering
\resizebox{0.70\linewidth}{!}{%
\begin{tikzpicture}[
    font=\small,
    neuron/.style={circle, draw=black, thick, minimum size=11mm},
    inp/.style={circle, draw=black, thick, minimum size=7mm},
    bias/.style={circle, draw=black, thick, minimum size=7mm},
    conn/.style={thick}
]

\node[inp] (x1) at (0,  1.2) {};
\node[inp] (x2) at (0, -1.2) {};
\node[bias] (x0) at (0, -3.0) {};
\node[left=6pt of x1] {$x_1$};
\node[left=6pt of x2] {$x_2$};
\node[left=6pt of x0] {$1$};

\node[neuron] (h1) at (3,  1.8) {};
\node[neuron] (h2) at (3,  0.0) {};
\node[neuron] (h3) at (3, -1.8) {};
\node[below=2pt of h3] {\scriptsize hidden layer ($h$ units)};

\node[neuron] (y) at (6, 0) {};
\node[right=8pt of y] {$z(x)$};

\draw[conn] (x1) -- (h1);
\draw[conn] (x2) -- (h1);
\draw[conn] (x0) -- (h1);

\draw[conn] (x1) -- (h2);
\draw[conn] (x2) -- (h2);
\draw[conn] (x0) -- (h2);

\draw[conn] (x1) -- (h3);
\draw[conn] (x2) -- (h3);
\draw[conn] (x0) -- (h3);

\draw[conn] (h1) -- (y);
\draw[conn] (h2) -- (y);
\draw[conn] (h3) -- (y);

\node[bias] (b2) at (6, -2.2) {};
\node[below=2pt of b2] {\scriptsize $1$};
\draw[conn] (b2) -- (y);

\node[align=center] at (3, 3.0) {\scriptsize $a=\phi(W_1x+b_1)$};
\node[align=center] at (6, 1.6) {\scriptsize $z=W_2a+b_2$};

\end{tikzpicture}
}
\caption{Schematic illustration of a one-hidden-layer multilayer perceptron (MLP). 
Only a subset of hidden units is shown; the total number of hidden units $h$ is a tunable hyperparameter.}

\label{fig:mlp_arch}
\end{figure}
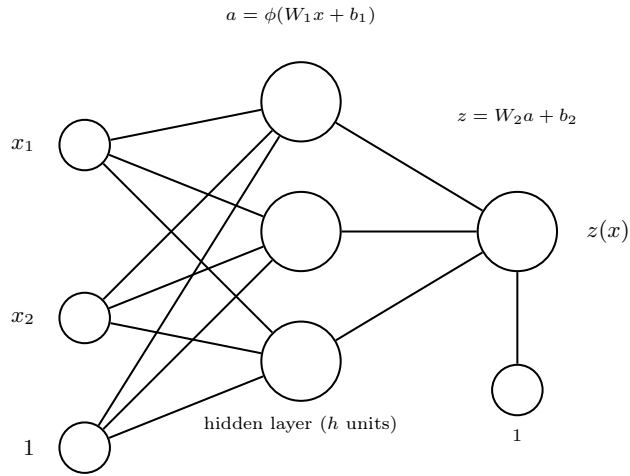
}

\FloatBarrier

\subsection{Quantum machine learning}
\label{subsec:quantum}

In this section, we briefly introduce the basic concepts of quantum machine learning required to understand the variational quantum classifier used in this work.
A comprehensive introduction to this field is provided by Schuld \emph{et al.}~\cite{schuld2014}.

\subsubsection{Qubits, superposition, and measurement}
\label{subsubsec:qubits}

The fundamental unit of quantum information is the \emph{qubit}. 
Mathematically, a qubit is represented as a normalized vector in a two-dimensional complex Hilbert space and can be written as
\begin{equation}
\ket{\psi} = \alpha \ket{0} + \beta \ket{1},
\label{eq:qubit}
\end{equation}
where $\ket{0}$ and $\ket{1}$ denote the computational basis states, and 
$\alpha, \beta \in \mathbb{C}$ are complex probability amplitudes satisfying the normalization condition
$|\alpha|^2 + |\beta|^2 = 1$.

When a qubit is measured in the computational basis $\{\ket{0}, \ket{1}\}$, the measurement outcomes are probabilistic and follow the Born rule:
\begin{equation}
\Pr(0) = |\alpha|^2, \qquad \Pr(1) = |\beta|^2.
\label{eq:born}
\end{equation}
Therefore, the squared magnitudes of the amplitudes determine the probabilities of observing each classical outcome.

\begin{figure}[h]
\centering
\includegraphics[width=0.45\linewidth]{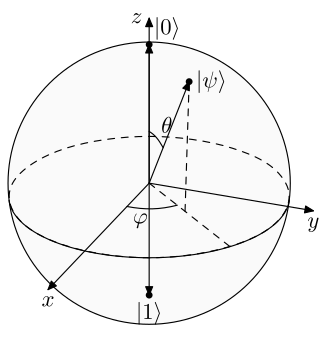}
\caption{Bloch sphere representation of a single qubit.
Any pure qubit state can be visualized as a point on the unit sphere,
parameterized by the polar and azimuthal angles $(\theta,\phi)$. 
\hfill {\small \textbf{Source:} Wikipedia -- Bloch sphere~\cite{wiki_bloch}.}}
\label{fig:bloch_sphere}
\end{figure}

For a system of $n$ qubits, the dimension of the state space grows exponentially as $2^n$. 
This exponential scaling enables uniquely quantum phenomena such as entanglement and interference, which form the basis of quantum information processing and quantum machine learning.

\noindent
The concepts of quantum bits, superposition, and measurement used in this work follow the standard formulation presented in Sec.~1.2 of
Nielsen and Chuang~\cite{nielsen2010}.

\subsubsection{Variational circuits and hybrid optimization}
\label{subsubsec:vqa}

Current quantum computers are still limited by noise and by the small number of available qubits.
Because of these constraints, many quantum machine learning models are based on
\emph{variational quantum algorithms} (VQAs), also known as variational hybrid
quantum--classical algorithms~\cite{mcclean2016theory}, which combine a quantum
circuit with a classical optimization loop.

In a VQA, the quantum part is a \emph{parameterized quantum circuit}.
This means that the circuit contains adjustable parameters, similar to weights in a neural network.
Mathematically, such a circuit implements a unitary transformation denoted by $U(\theta)$, where
\[
\theta = (\theta_1, \theta_2, \dots, \theta_P) \in \mathbb{R}^P
\]
is a vector of $P$ real-valued parameters.
Each parameter $\theta_k$ typically corresponds to a rotation angle of a quantum gate.

The circuit starts from a fixed and simple initial quantum state, usually chosen as
$\ket{0}^{\otimes n}$.
Here, $\ket{0}$ denotes the ground state of a single qubit, and the tensor power
$\ket{0}^{\otimes n}$ represents $n$ qubits all initialized in the state $\ket{0}$.

After applying the parameterized circuit $U(\theta)$, the quantum state becomes
\begin{equation}
\ket{\psi(\theta)} = U(\theta)\ket{0}^{\otimes n}.
\end{equation}
The notation $\ket{\psi(\theta)}$ emphasizes that the resulting quantum state depends on the parameters $\theta$.

The goal of training is to find parameter values $\theta$ that minimize a real-valued
\emph{cost function} $\mathcal{C}(\theta)$.
This cost function reflects how
well the quantum circuit performs a given task, such as classification.
Since quantum states cannot be directly accessed, the value of $\mathcal{C}(\theta)$ is estimated from repeated quantum measurements.
A classical optimizer then updates the parameters $\theta$ based on these measurement results.

Increasing the depth of the circuit (i.e., using more parameterized gates) generally makes the model more expressive.
However, deep variational circuits may suffer from optimization problems known as
\emph{barren plateaus}~\cite{cerezo2021costdependent,larocca2025barren}, where the gradients of the cost function become extremely small.
In this regime, training becomes difficult because parameter updates provide almost no improvement.

\subsubsection{Feature map: encoding classical data into quantum states}
\label{subsubsec:featuremap}

Quantum circuits operate on quantum states, whereas real-world data are classical.
Therefore, classical input data must first be encoded into a quantum state.
Let
\[
x = (x_1, x_2, \dots, x_d) \in \mathbb{R}^d
\]
be a classical data point with $d$ real-valued features.

The encoding is performed by an input-dependent unitary transformation, called a

\emph{feature map}, and denoted by $U_\phi(x)$.
Applying this transformation to the initial state $\ket{0}^{\otimes n}$ produces the quantum state
\begin{equation}
\ket{\phi(x)} = U_\phi(x)\ket{0}^{\otimes n}.
\label{eq:featuremap}
\end{equation}

Here, the symbol $\phi$ indicates that the classical data $x$ are mapped to a quantum feature representation.
In practical implementations, the individual
components of $x$ are often encoded directly as rotation angles of single-qubit
gates, which provides a simple and transparent encoding scheme.

The feature map embeds the classical data into a high-dimensional quantum Hilbert space.
This embedding is typically nonlinear, meaning that simple classical data can lead to complex quantum states.
From a machine learning perspective, this behavior is closely related to kernel methods:
inner products and expectation values of quantum states implicitly define similarity measures between data points in the quantum feature space~\cite{havlicek2019quantumkernel}. General strategies for encoding classical data into quantum states are reviewed
in~\cite{schuld2018supervised}.

\subsubsection{Variational Quantum Classifier (VQC)}
\label{subsubsec:vqc}

A variational quantum classifier (VQC) combines data encoding and trainable quantum processing.
It consists of two consecutive circuits:
\begin{itemize}
\item a feature map $U_\phi(x)$ that encodes the classical input data,
\item a trainable variational circuit (ansatz) $U(\theta)$ that processes the encoded data.
\end{itemize}

Variational quantum classifiers are a standard instance of supervised quantum
learning models~\cite{schuld2014,schuld2018supervised}.

We denote by $L$ the ansatz depth, i.e., the number of repeated trainable layers.
Accordingly, the trainable unitary can be written as
\begin{equation}
U(\theta) = \prod_{\ell=1}^{L} U_\ell(\theta_\ell),
\label{eq:ansatz_depth}
\end{equation}
where $U_\ell(\theta_\ell)$ denotes the $\ell$-th variational layer and
$\theta = \{\theta_\ell\}_{\ell=1}^{L}$ collects all trainable parameters.

After applying both circuits, the classification result is obtained by measuring a quantum observable.
We denote this observable by $M$.
In most applications, $M$ is chosen as a Pauli operator, such as the Pauli-$Z$ operator acting on one qubit.

The model output is defined as the expectation value of $M$ in the final quantum state:
\begin{equation}
f_\theta(x) =
\bra{0}^{\otimes n}
U_\phi^\dagger(x)\, U^\dagger(\theta)\, M\, U(\theta)\, U_\phi(x)
\ket{0}^{\otimes n}.
\label{eq:vqc_expect}
\end{equation}

This expression can be read as follows.
The circuit $U_\phi(x)$ encodes the input data, the circuit $U(\theta)$ processes it,
and the observable $M$ is measured.

The dagger symbol $(\cdot)^\dagger$ denotes the Hermitian adjoint and ensures that the expression corresponds to an expectation value.

If $M$ is chosen as the Pauli-$Z$ operator, the output satisfies
\[
f_\theta(x) \in [-1, 1].
\]
To interpret this value as a probability for binary classification, we adopt the convention
that $f_\theta(x)=+1$ corresponds to class $0$ and $f_\theta(x)=-1$ to class $1$.
Accordingly, the output is mapped to a probability as
\begin{equation}
p_\theta(y=1 \mid x) = \frac{1 - f_\theta(x)}{2}.
\label{eq:vqc_prob}
\end{equation}

This mapping ensures that $p_\theta(y=1 \mid x) \in [0,1]$.

The parameters $\theta$ are trained by minimizing the binary cross-entropy loss over a labeled dataset
$\{(x_i, y_i)\}_{i=1}^N$, where $y_i \in \{0,1\}$ are the class labels:
\begin{equation}
\mathcal{L}(\theta) =
-\frac{1}{N}\sum_{i=1}^N
\left[
y_i \log p_\theta(y=1 \mid x_i)
+ (1 - y_i)\log\!\left(1 - p_\theta(y=1 \mid x_i)\right)
\right].
\label{eq:vqc_ce}
\end{equation}

\paragraph{Expressivity, depth, and measurement noise.}
The ansatz depth $L$ controls the expressive capacity of the variational circuit
by determining how many layers of parameterized gates and entangling operations
are applied.
While increasing $L$ generally allows the circuit to represent more complex
decision functions, it also leads to a larger parameter space and can make the
optimization landscape more challenging, particularly in the presence of noise.

On real quantum hardware, expectation values are estimated from a finite number
of measurement samples, referred to as \emph{shots}.
Finite-shot estimation introduces statistical fluctuations in the measured
expectation values, such that repeated evaluations of the same circuit with
fixed parameters $\theta$ may yield different outcomes.
This sampling noise affects both the model output and the gradient estimates
used during training, and plays an important role in the practical performance
of variational quantum classifiers.

\begin{figure}[h]
\centering
\begin{quantikz}
\lstick{$\ket{0}$} & \gate{U_\phi(x)} & \gate{U(\theta)} & \meter{$Z$} \\
\lstick{$\ket{0}$} & \gate{U_\phi(x)} & \gate{U(\theta)} & \qw
\end{quantikz}
\caption{High-level structure of a Variational Quantum Classifier (VQC): feature map, trainable ansatz, and measurement.}
\label{fig:vqc_scheme}
\end{figure}
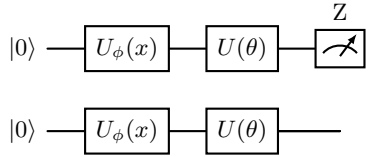

\newpage
\section{Problem Formulation: XOR Classification}
\label{sec:problem}

In this work, we study the binary classification problem known as the
\emph{exclusive OR} (XOR).
Given an input vector
\[
x = (x_1, x_2)^\top \in \mathbb{R}^2,
\]
the target label is defined as
\[
y = \mathrm{XOR}(x_1, x_2) \in \{0,1\},
\]
where the XOR rule assigns label $1$ if exactly one of the two inputs is active
and label $0$ otherwise.

In its canonical discrete form, the XOR problem consists of four input--output
pairs:
\[
(0,0)\mapsto 0,\quad
(0,1)\mapsto 1,\quad
(1,0)\mapsto 1,\quad
(1,1)\mapsto 0.
\]
This configuration is a minimal example of a \emph{linearly inseparable}
classification task: no single linear decision boundary can separate the two
classes in the input space.
As shown in Section~\ref{subsubsec:xor_proof}, any linear classifier necessarily
fails on this dataset.

We note that, within the quantum approach, classical input features typically require appropriate scaling before encoding into quantum circuits. In particular, when embedding data using rotational gates, input values are typically mapped to a bounded interval, such as $[0,\pi]$ or $[0,2\pi]$, due to the periodic nature of quantum rotations \cite{schuld2018supervised}. Although the canonical XOR problem is defined on binary inputs, this scaling becomes relevant for continuous and noisy versions of the problem considered in quantm experiments.

To analyze model behavior beyond the idealized setting, we consider several
extensions of the XOR problem that introduce noise and continuous structure.
These variants allow us to systematically study expressivity, robustness to
noise, and generalization of classical and quantum models.

\paragraph{Evaluation metrics.}
For all problem variants, model performance is evaluated using:
\begin{itemize}
\item \textbf{Accuracy}: the fraction of correctly classified samples,
\item \textbf{Loss}: Binary cross-entropy (BCE),
      depending on the model and experiment.
\end{itemize}
Both training and test metrics are reported.

\begin{figure}[h]
\centering

\begin{subfigure}[t]{0.32\textwidth}
    \centering
    \includegraphics[width=\linewidth]{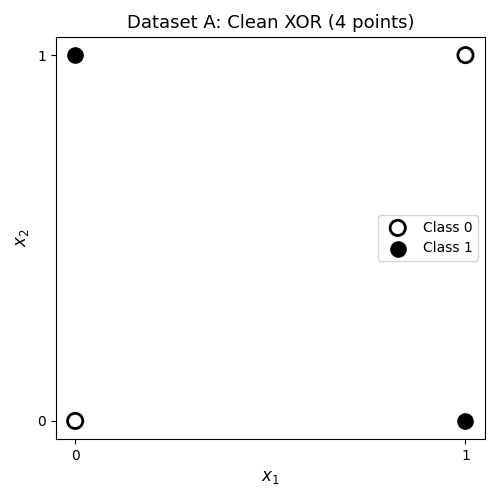}
    \label{fig:dataset_A}
\end{subfigure}
\hfill
\begin{subfigure}[t]{0.32\textwidth}
    \centering
    \includegraphics[width=\linewidth]{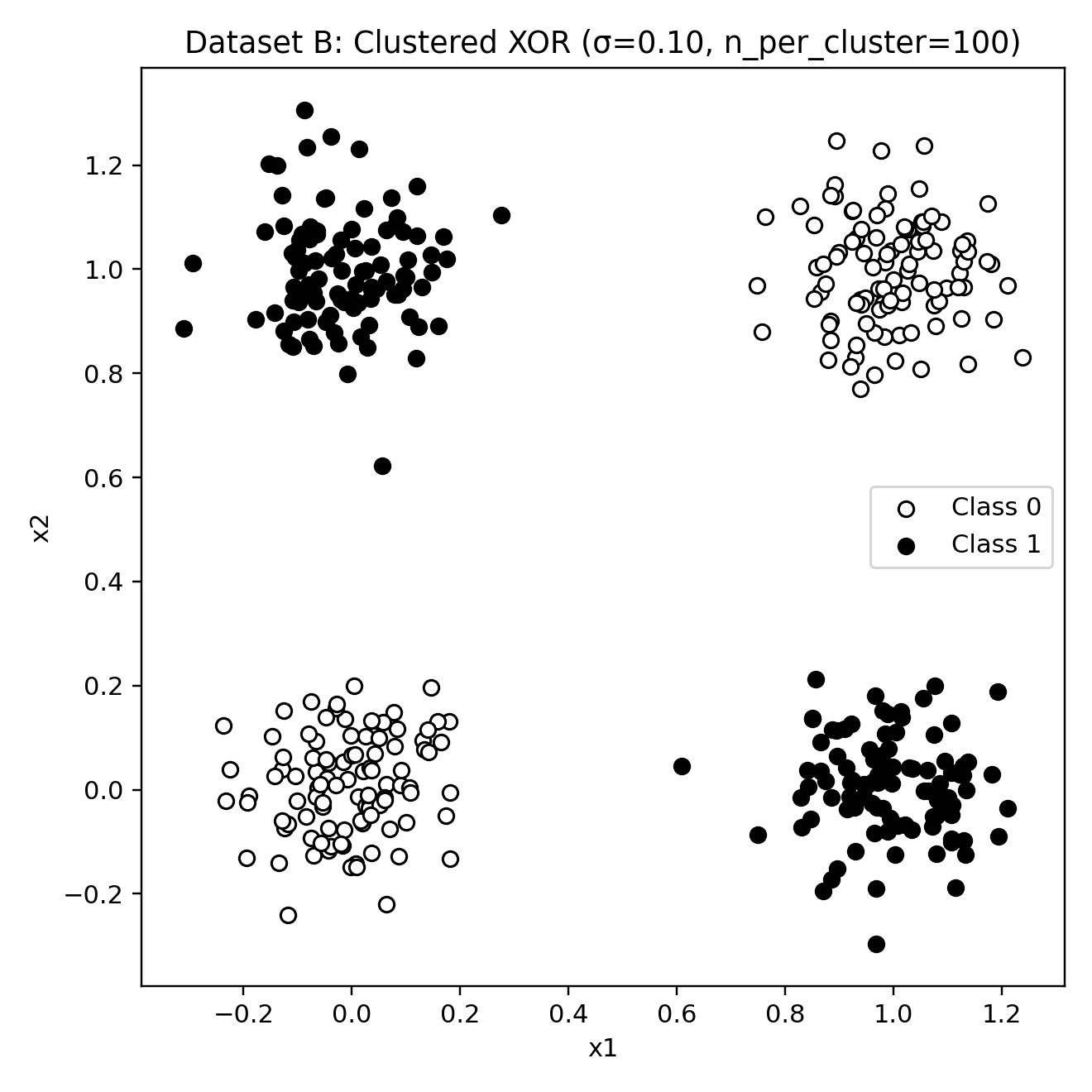}
    \label{fig:dataset_B}
\end{subfigure}
\hfill
\begin{subfigure}[t]{0.32\textwidth}
    \centering
    \includegraphics[width=\linewidth]{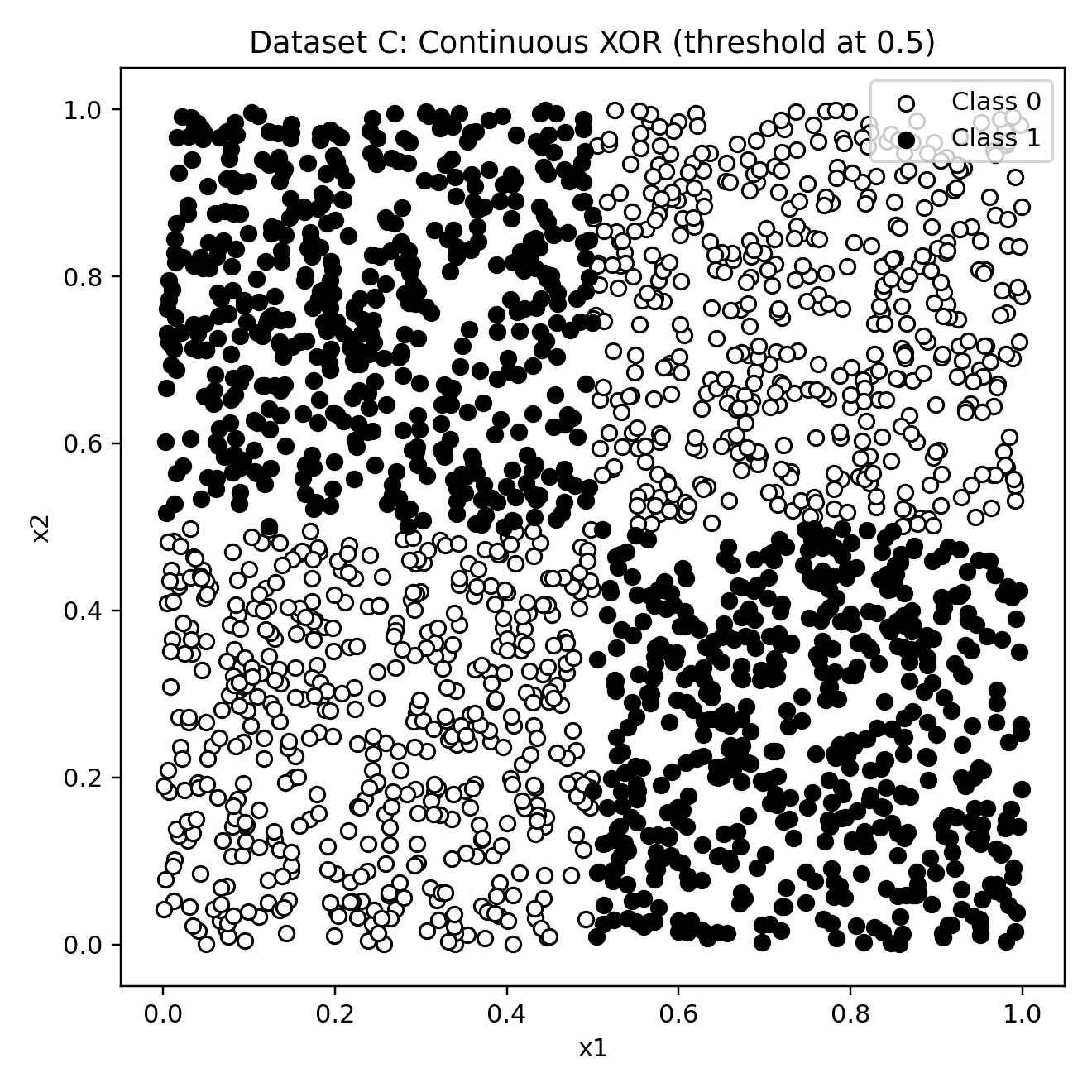}
    \label{fig:dataset_C}
\end{subfigure}

\caption{Illustration of the three XOR dataset variants used in this study:
(a) Dataset A — clean, discrete XOR;
(b) Dataset B — clustered XOR with additive Gaussian noise;
(c) Dataset C — continuous XOR defined by a threshold-based rule.}
\label{fig:xor_datasets}
\end{figure}

\newpage
\section{Methodology and Experimental Setup}
\label{sec:methodology}

This section describes the datasets, experimental protocol, and model
configurations used to compare classical and quantum approaches to XOR
classification.

\subsection*{Datasets}

We consider three synthetic datasets of increasing complexity.

\paragraph{Dataset A: Clean XOR.}
This dataset consists of the four input points
$(0,0)$, $(0,1)$, $(1,0)$, and $(1,1)$, with class labels assigned according to
the XOR rule.
\paragraph{Dataset B: Noisy Clustered XOR.}
This dataset is the main benchmark used in the experiments.
Four clusters are generated around the same XOR corner points, but with
additive Gaussian noise.
The noise level is controlled by the standard deviation $\sigma$.

We vary:
\[
\sigma \in \{0.00, 0.05, 0.10, 0.20, 0.30\},
\qquad
n_{\text{per cluster}} \in \{25, 50, 100, 250, 500\}.
\]

\paragraph{Dataset C: Continuous XOR (threshold-based).}
Points are sampled uniformly from the unit square $[0,1]^2$.
Class labels are assigned using a threshold $t$:
a point belongs to class $1$ if exactly one coordinate exceeds $t$.
This dataset provides a smoother, continuous version of XOR and is used mainly
for qualitative analysis and discussion.
Since the primary focus of this work is on robustness and comparative benchmarking,
detailed results for Dataset~C are deferred to Appendix~\ref{app:datasetC}.

\subsection*{Experimental Protocol}

All datasets are split into train/test sets (80/20) using a fixed data seed
($\texttt{seed}=42$) for generation and splitting.

To account for randomness in model initialization and optimization, each
experiment is repeated with five different model seeds:
\[
\texttt{seeds} \in \{0,1,2,3,4\}.
\]
Unless stated otherwise, reported results are given as mean $\pm$ standard
deviation over these runs.

For seed-sensitivity analyses, we use an extended set $\texttt{seeds}=0,\dots,19$;
all other results use $\texttt{seeds}=0,\dots,4$ for consistency.

The following quantities are recorded for each model and dataset configuration:
\begin{itemize}
\item training and test accuracy,
\item training and test loss,
\item number of trainable parameters,
\item wall-clock training time.
\end{itemize}
All dataset parameters and model configurations are fixed and summarized in
Tables~\ref{tab:exp_settings_data} and~\ref{tab:exp_settings_models}.

{
\setlength{\tabcolsep}{10pt}
\renewcommand{\arraystretch}{0.95}
\begin{table}[h]
\centering
\footnotesize
\begin{tabularx}{\textwidth}{p{3.0cm} X}
\toprule
\textbf{Parameter} & \textbf{Value} \\
\midrule

\multicolumn{2}{l}{\textbf{Datasets}} \\
Types &
Clean XOR (A), Noisy Clustered XOR (B), Continuous XOR (C) \\
Noise level (Dataset B) &
$\sigma \in \{0.00, 0.05, 0.10, 0.20, 0.30\}$ \\
Samples per cluster (Dataset B) &
$\{25, 50, 100, 250, 500\}$ \\
Threshold (Dataset C) &
$t = 0.5$ \\

\midrule
\multicolumn{2}{l}{\textbf{Training protocol}} \\
Train / test split &
80\% / 20\% \\
Repetitions &
5 runs per setting (mean $\pm$ std over different seeds) \\

\midrule
\multicolumn{2}{l}{\textbf{Random seeds}} \\
Dataset seed & 42 \\
Model seeds & $\{0,1,2,3,4\}$ \\

\bottomrule
\end{tabularx}
\caption{Experimental settings: datasets, training protocol, and random seeds.}
\label{tab:exp_settings_data}
\end{table}
}

{
\setlength{\tabcolsep}{10pt}
\renewcommand{\arraystretch}{0.95}
\begin{table}[!htbp]
\centering
\footnotesize
\begin{tabularx}{\textwidth}{p{3.0cm} X}
\toprule
\textbf{Parameter} & \textbf{Value} \\
\midrule

\multicolumn{2}{l}{\textbf{Classical models}} \\
Logistic regression &
2 input features + bias (3 trainable parameters) \\
MLP hidden units &
$h \in \{1,2,4,8\}$ \\
Activation function &
Sigmoid \\
Optimizer &
Stochastic Gradient Descent (SGD) \\
Training epochs & 3000--5000 (depending on setting) \\

\midrule
\multicolumn{2}{l}{\textbf{Quantum model}} \\
Number of qubits & 2 \\
Encoding &
Angle encoding: each feature is mapped to a rotation angle of a single-qubit gate \\
Ansatz depth &
$L \in \{1,2\}$ \\
Trainable parameters &
$6L$ \\
Observable &
Pauli-$Z$ expectation value \\
Optimizer &
Adam / Gradient Descent \\
Learning rate &
$lr \in [0.1, 0.3]$ (fixed after pilot study) \\
Measurement shots &
analytic (shot-free), $\{128,1024\}$ \\

\bottomrule
\end{tabularx}
\caption{Model configurations used in this study (classical and quantum).}
\label{tab:exp_settings_models}
\end{table}
}

\FloatBarrier

\subsection*{Classical Models}

\paragraph{Linear classifier.}
As a baseline, we use logistic regression with two input features and a bias
term.
The model has three trainable parameters.
Due to the linear inseparability of XOR, its expected performance is close to
random guessing on clean and noisy XOR datasets.

\paragraph{Multilayer Perceptron (MLP).}
We employ a fully connected neural network with a single hidden layer.
The hidden-layer width is varied as
\[
h \in \{1, 2, 4, 8\}.
\]
In all main experiments comparing classical and quantum models, the MLP
architecture is fixed to \(h=4\) hidden units.
Both the hidden and output layers use sigmoid activation functions, with
nonlinearity introduced at the hidden layer.
Training is performed using stochastic gradient descent for a fixed number of
epochs.
The MLP serves as a strong classical nonlinear baseline.

To assess the influence of model capacity, we conduct a separate ablation study
in which the number of hidden units \(h\) is varied while all other
hyperparameters are kept fixed.
The results of this analysis are reported in Appendix~\ref{app:mlp_width_ablation} and
motivate the choice of \(h=4\) as a compact and sufficient baseline for the main
quantum--classical comparison.

\subsection*{Quantum Model}

\paragraph{Variational Quantum Classifier (VQC).}
The quantum model is implemented as a variational quantum circuit with two
qubits.
Classical inputs are encoded using angle encoding:
\[
x_1 \mapsto R_X(\pi x_1), \qquad
x_2 \mapsto R_X(\pi x_2).
\]

The variational ansatz consists of $L$ repeated layers, with
\[
L \in \{1,2\}.
\]
Each layer applies parameterized single-qubit rotations followed by an
entangling CNOT gate.
The number of trainable parameters is $6L$.

Measurement is performed on one qubit using the Pauli-$Z$ operator.
We denote the expectation value by
\[
m = \langle Z \rangle = f_\theta(x) \in [-1,1].
\]

We adopt the convention that
\[
m = +1 \;\rightarrow\; y=1,
\qquad
m = -1 \;\rightarrow\; y=0.
\]

To obtain a probabilistic interpretation, we linearly rescale
\[
p_\theta(y=1 \mid x) = \frac{1 + m}{2}.
\]

Equivalently,
\[
m = 2\,p_\theta(y=1 \mid x) - 1.
\]

Training minimizes the binary cross-entropy loss using a classical optimizer.
Experiments are performed both in the analytic (shot-free) regime and with
finite sampling noise using
\[
\text{shots} \in \{128, 1024\}.
\]

A structured summary of all VQC configurations used in the experiments,
including circuit depth, number of parameters, shot settings, encoding, and
backend details, is provided in Table~\ref{tab:vqc_settings}.

\begin{table}[!h]
\caption{Variational quantum classifier (VQC) configurations used in the experiments.}
\label{tab:vqc_settings}
\centering
\small
\setlength{\tabcolsep}{5pt}
\begin{tabular}{c c c l l l}
\hline
$L$ & \#params & shots & encoding & observable & device/interface \\
\hline
1 & 6  & analytic & $RX(\pi x_1), RX(\pi x_2)$ & $\langle Z_0 \rangle$ & default.qubit / autograd \\
1 & 6  & 128      & $RX(\pi x_1), RX(\pi x_2)$ & $\langle Z_0 \rangle$ & default.qubit / autograd \\
1 & 6  & 1024     & $RX(\pi x_1), RX(\pi x_2)$ & $\langle Z_0 \rangle$ & default.qubit / autograd \\
2 & 12 & analytic & $RX(\pi x_1), RX(\pi x_2)$ & $\langle Z_0 \rangle$ & default.qubit / autograd \\
2 & 12 & 1024     & $RX(\pi x_1), RX(\pi x_2)$ & $\langle Z_0 \rangle$ & default.qubit / autograd \\
\hline
\end{tabular}
\end{table}

\subsection*{Analysis and Visualization}

To support quantitative and qualitative comparison, we report:
\begin{itemize}
\item decision boundaries for representative settings,
\item learning curves (loss vs.\ training iteration),
\item accuracy as a function of noise level $\sigma$,
\item accuracy as a function of dataset size,
\item accuracy as a function of the number of shots for the VQC.
\end{itemize}

\section{Experiments and Results}
\label{sec:experiments_results}

This section presents a comparative experimental evaluation of classical and
quantum variational classifiers on the XOR problem under a unified experimental
protocol.
We focus on qualitative and quantitative aspects of model behavior, including
learned decision boundaries, training dynamics, robustness to noise and dataset
size, sensitivity to random initialization, and summary statistics such as model
size, training time, and predictive accuracy.

\subsection{Decision Boundaries}
\label{subsec:db_all}

We begin by comparing the decision boundaries obtained by classical and quantum classifiers on representative XOR problem configurations. From the figures we can see how each model partitions the two-dimensional input space after training, as a result revealing us the geometric structure of the resulting decision function and the model's ability to represent nonlinear class separations.

\begin{figure}[h]
\centering
\captionsetup[subfigure]{font=footnotesize,skip=1pt}

\begin{subfigure}[t]{0.32\textwidth}
  \centering
  \includegraphics[width=\linewidth]{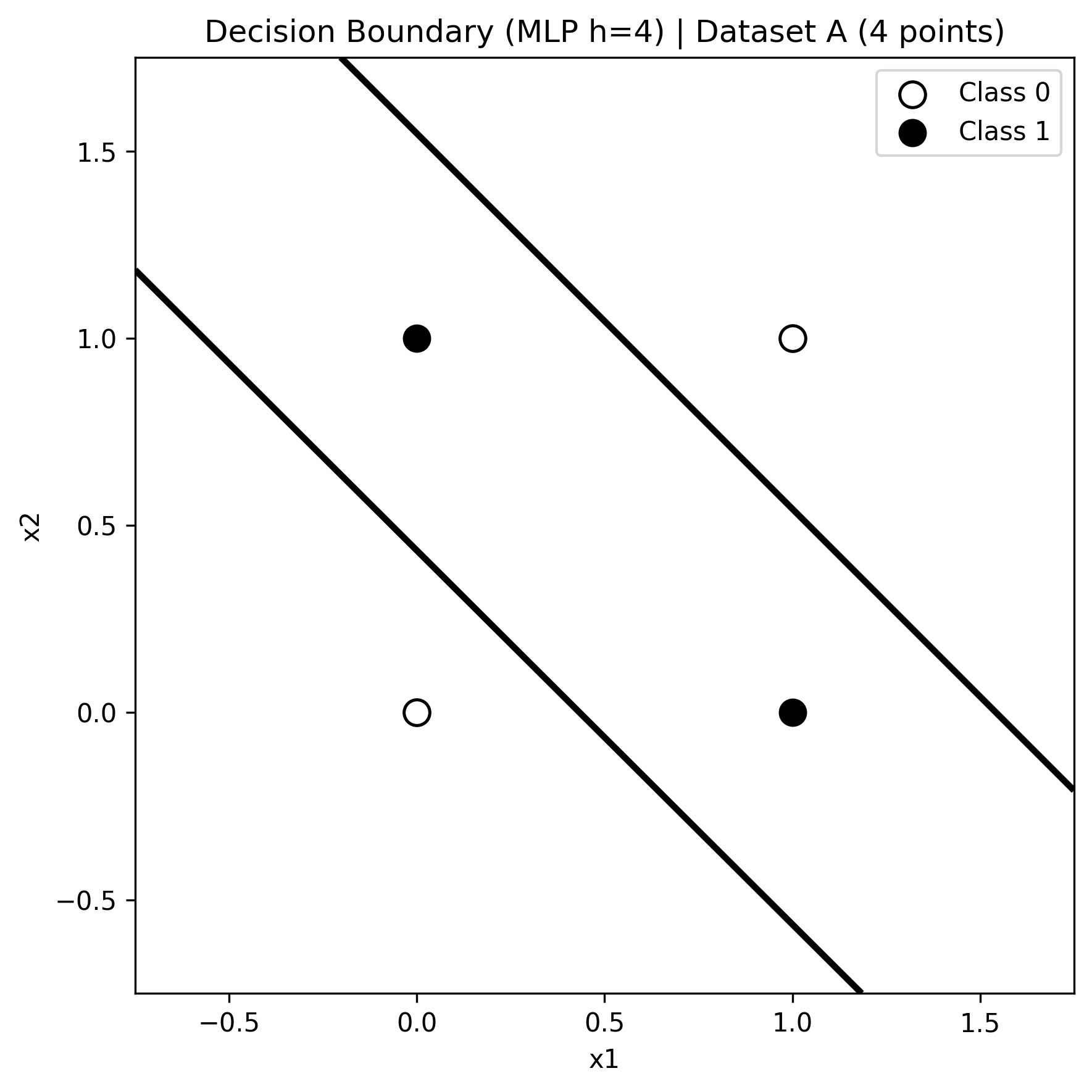}
  \caption{MLP: Dataset A}
  \label{fig:db_mlp_A}
\end{subfigure}
\hfill
\begin{subfigure}[t]{0.32\textwidth}
  \centering
  \includegraphics[width=\linewidth]{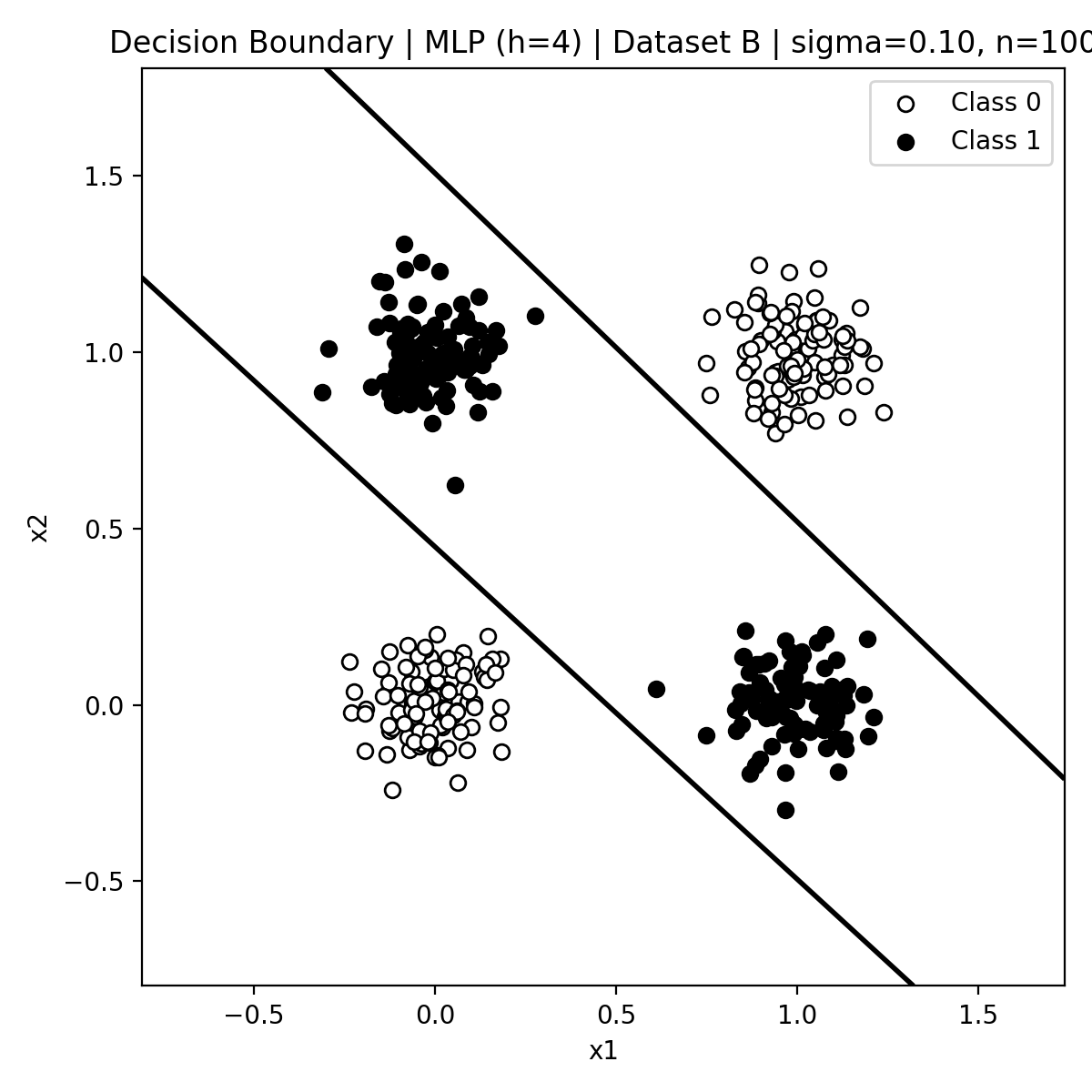}
  \caption{MLP: B ($\sigma=0.10$)}
  \label{fig:db_mlp_B010}
\end{subfigure}
\hfill
\begin{subfigure}[t]{0.32\textwidth}
  \centering
  \includegraphics[width=\linewidth]{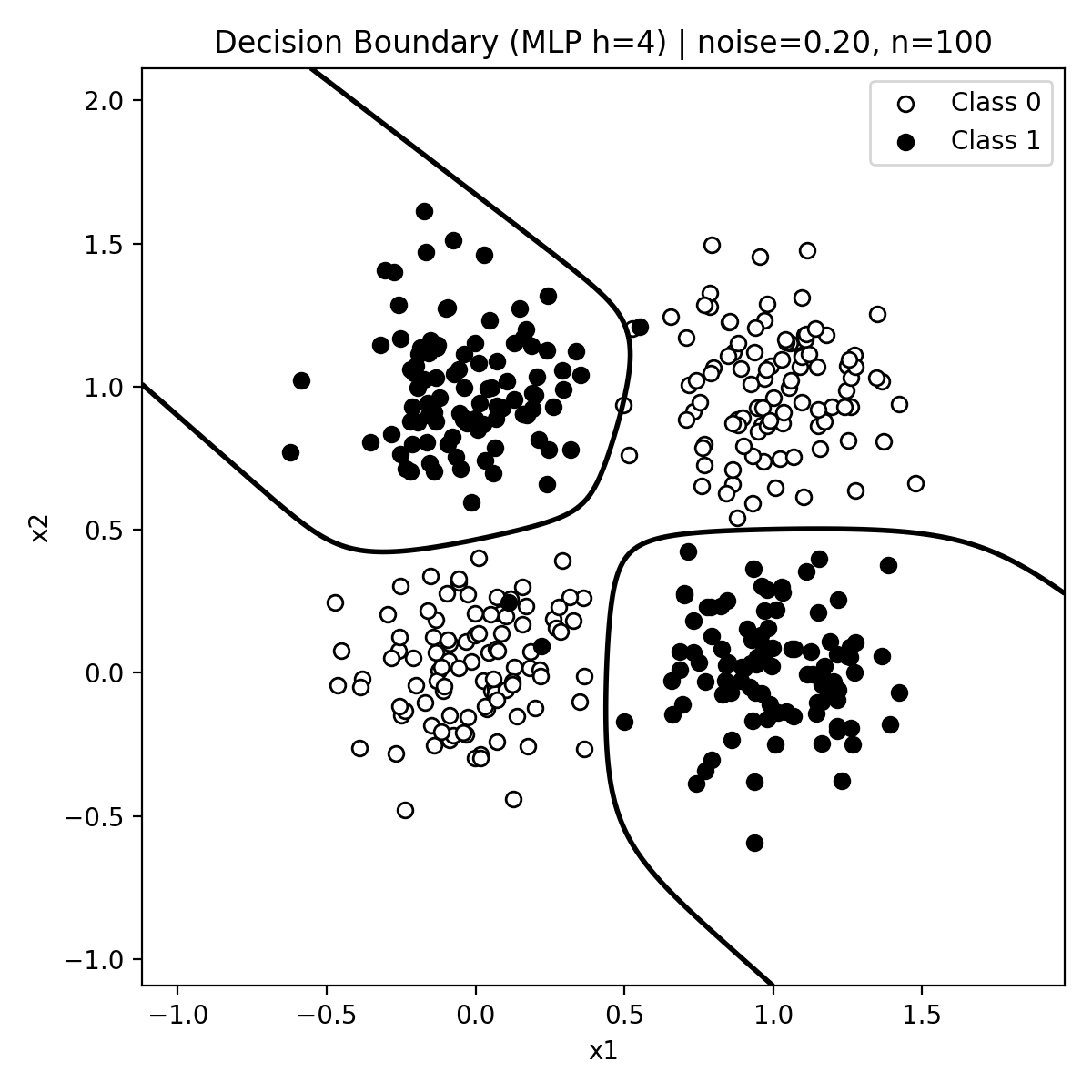}
  \caption{MLP: B ($\sigma=0.20$)}
  \label{fig:db_mlp_B020}
\end{subfigure}

\caption{Decision boundaries learned by the classical MLP ($h=4$) on representative XOR settings.}
\label{fig:db_mlp_group}
\end{figure}

Figures~\ref{fig:db_mlp_group} and~\ref{fig:db_vqc_group_2x2} present a qualitative comparison of the decision boundaries obtained by classical and quantum models on representative XOR configurations. The plots visualize the segmentation of the two-dimensional input space obtained by each model after training. 

For the classical MLP, Figure ~\ref{fig:db_mlp_group} shows the resulting decision boundaries for the pure XOR dataset (Dataset A) and for the clustered noisy XOR dataset (Dataset B) at two different noise levels. In the noise-free case, the MLP generates a nonlinear decision boundary that correctly separates the four XOR regions in the input space. With small amount of noise ($\sigma=0.10$), the overall XOR structure remains visible, while the boundary adapts to the clustered distribution of the data.

However, at higher noise levels (e.g., $\sigma=0.20$), the boundary becomes less clearly aligned with the ideal XOR structure, representing increased overlap between clusters while maintaining its characteristic global XOR geometry.

\begin{figure}[!t]
\centering
\captionsetup[subfigure]{font=footnotesize,skip=1pt}

\begin{subfigure}[t]{0.48\textwidth}
  \centering
  \includegraphics[width=\linewidth]{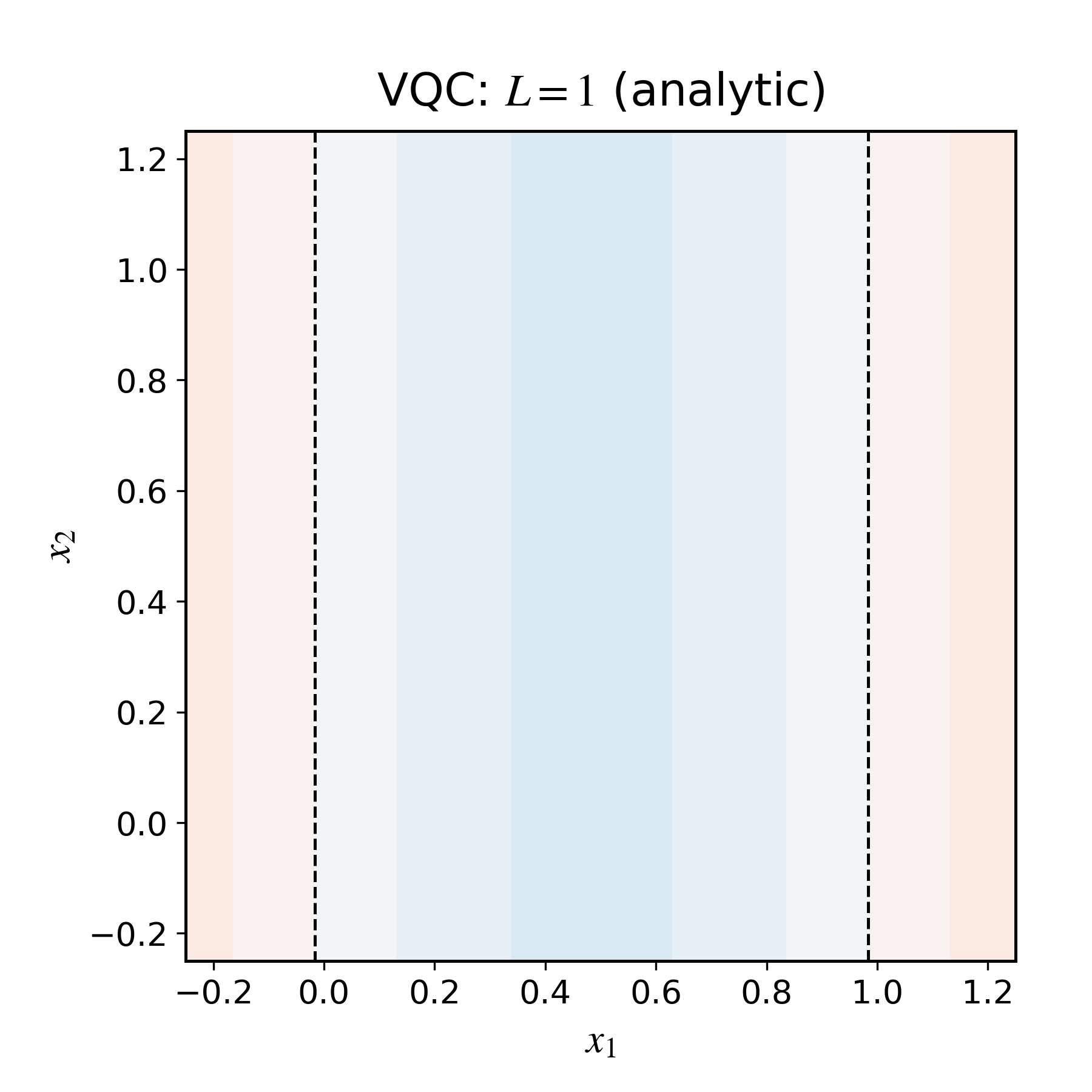}
  \caption{VQC: $L=1$ (analytic)}
  \label{fig:db_vqc_L1_analytic}
\end{subfigure}
\hfill
\begin{subfigure}[t]{0.48\textwidth}
  \centering
  \includegraphics[width=\linewidth]{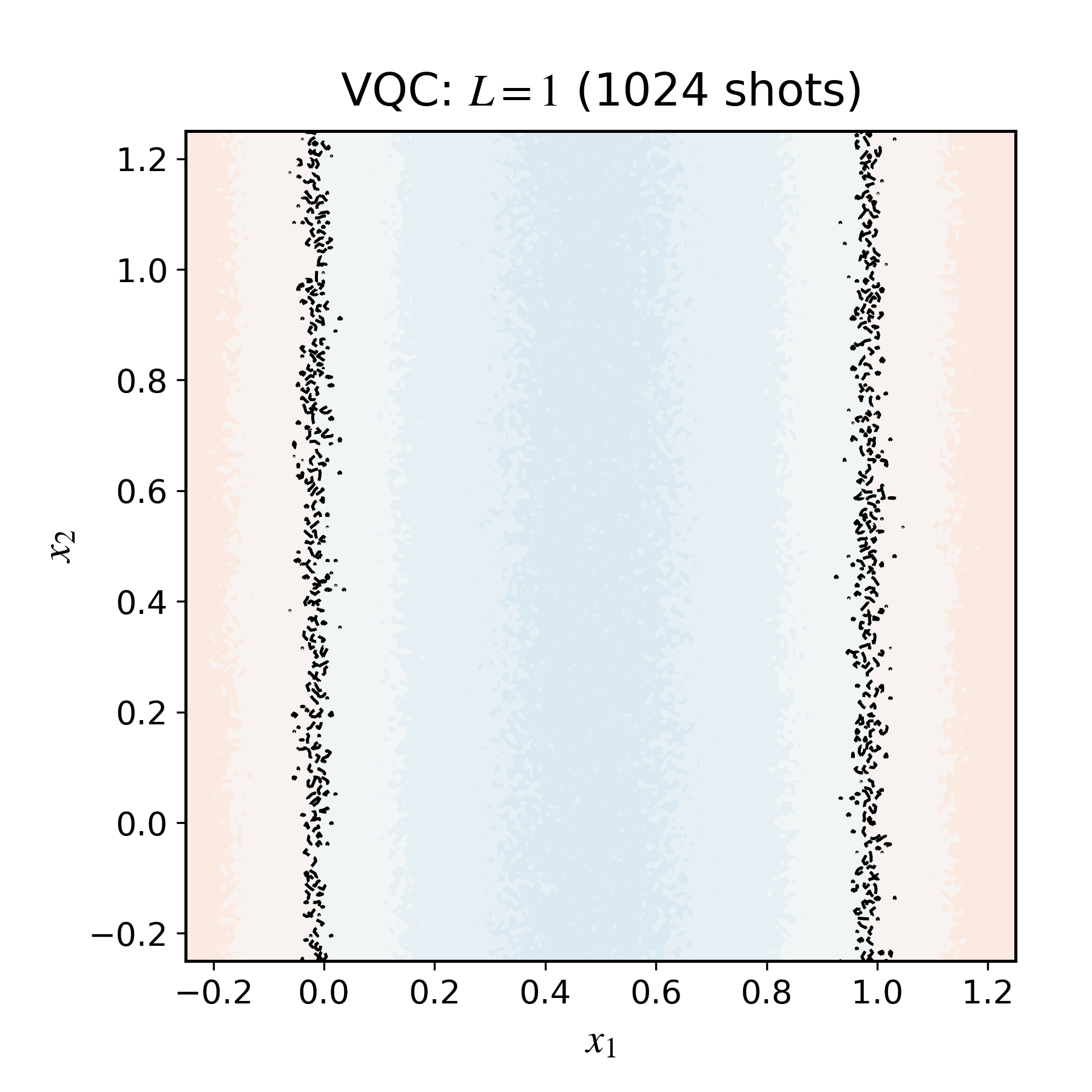}
  \caption{VQC: $L=1$ (1024 shots)}
  \label{fig:db_vqc_L1_1024}
\end{subfigure}

\vspace{0.35em}

\begin{subfigure}[t]{0.48\textwidth}
  \centering
  \includegraphics[width=\linewidth]{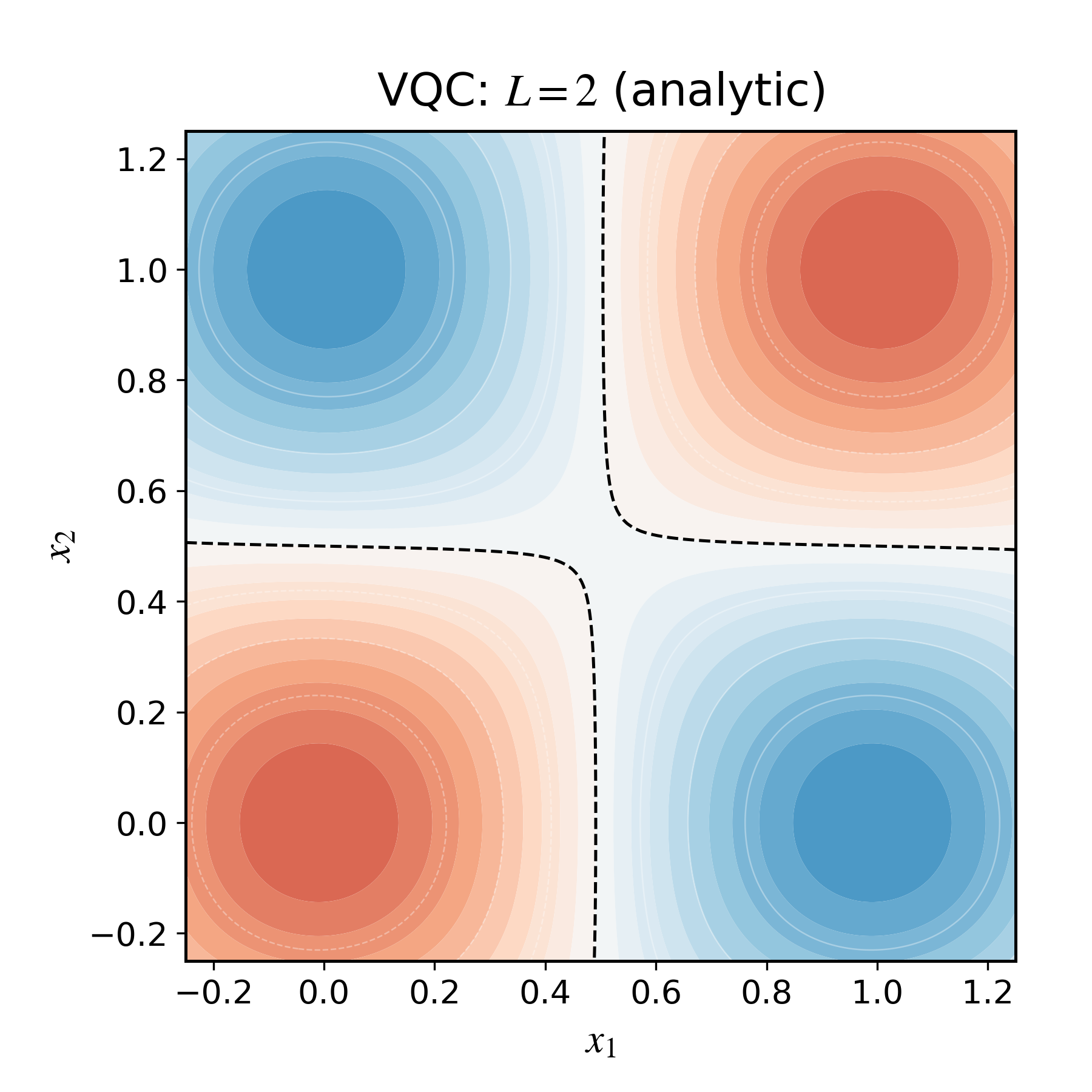}
  \caption{VQC: $L=2$ (analytic)}
  \label{fig:db_vqc_L2_analytic}
\end{subfigure}
\hfill
\begin{subfigure}[t]{0.48\textwidth}
  \centering
  % TODO: you will add this file
  \includegraphics[width=\linewidth]{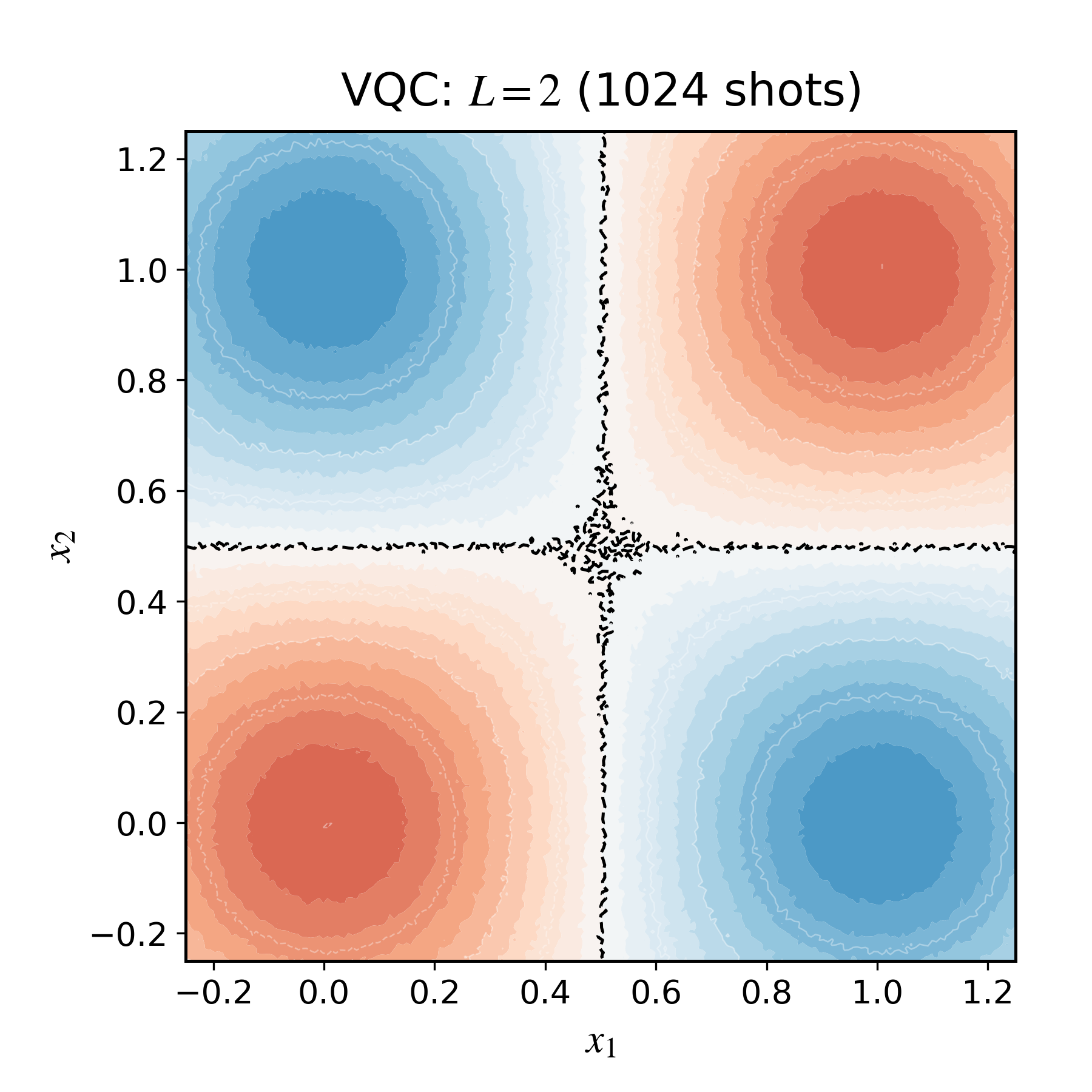}
  \caption{VQC: $L=2$ (1024 shots)}
  \label{fig:db_vqc_L2_1024}
\end{subfigure}

\vspace{0.45em}

\includegraphics[width=0.86\linewidth]{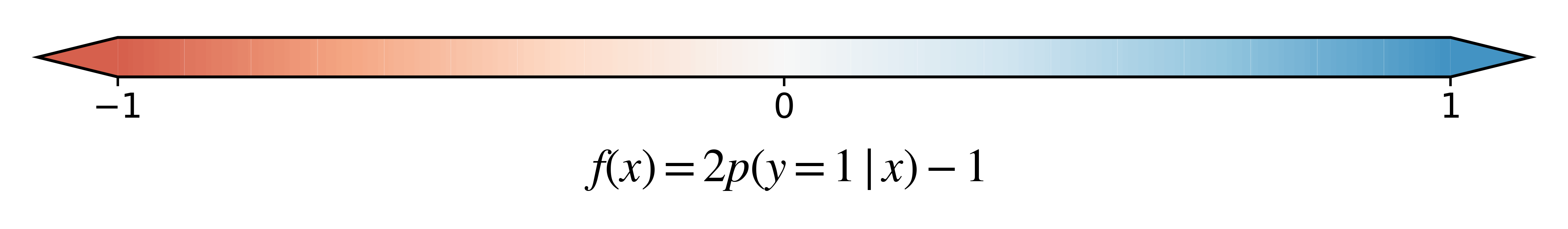}

\caption{Decision boundaries learned by the VQC.
The background color represents the prediction score
$f(x) = 2p(y=1\mid x)-1 \in [-1,1]$ using a shared color scale.
The dashed line indicates the decision boundary ($p=0.5$).}
\label{fig:db_vqc_group_2x2}
\end{figure}

Figure~\ref{fig:db_vqc_group_2x2} shows the decision boundaries learned by the
variational quantum classifier (VQC) for different circuit depths and measurement
regimes.
The background color, that you can see, encodes the continuous prediction score
$f(x)=2p(y=1\mid x)-1$, with the dashed curve indicating the decision boundary
corresponding to $p=0.5$.
For a shallow circuit ($L=1$) in the analytic setting, the VQC produces a simple
nonlinear partitioning of the input space that captures the XOR structure at a
coarse level.
When finite-shot measurements are used, additional small-scale oscillations
appear in the boundary, while the overall geometry remains comparable.

Increasing the circuit depth to $L=2$ leads to more structured decision regions
in the analytic regime, with a clearer separation of the all four XOR variants.
The corresponding finite-shot result exhibits visible sampling-induced
irregularities along the decision boundary, while maintaining the same global
partitioning pattern.
Across all panels, the VQC decision boundaries are shown under a shared color
scale, enabling direct visual comparison between circuit depths and measurement
regimes.

Overall, these figures illustrate how classical and quantum models partition the
input space for the XOR task under different conditions.
Quantitative comparisons and further analysis of learning dynamics, robustness,
and depth dependence are provided.

\subsection{Learning Behavior}
\label{subsec:learning_behavior}

Figure~\ref{fig:learning_behavior} presents representative learning curves for the
multilayer perceptron (MLP) and the variational quantum classifier (VQC), shown in
panels (a)--(f).

\begin{figure}[h]
\centering
\begin{subfigure}[t]{0.32\textwidth}
  \centering
  \includegraphics[width=\linewidth]{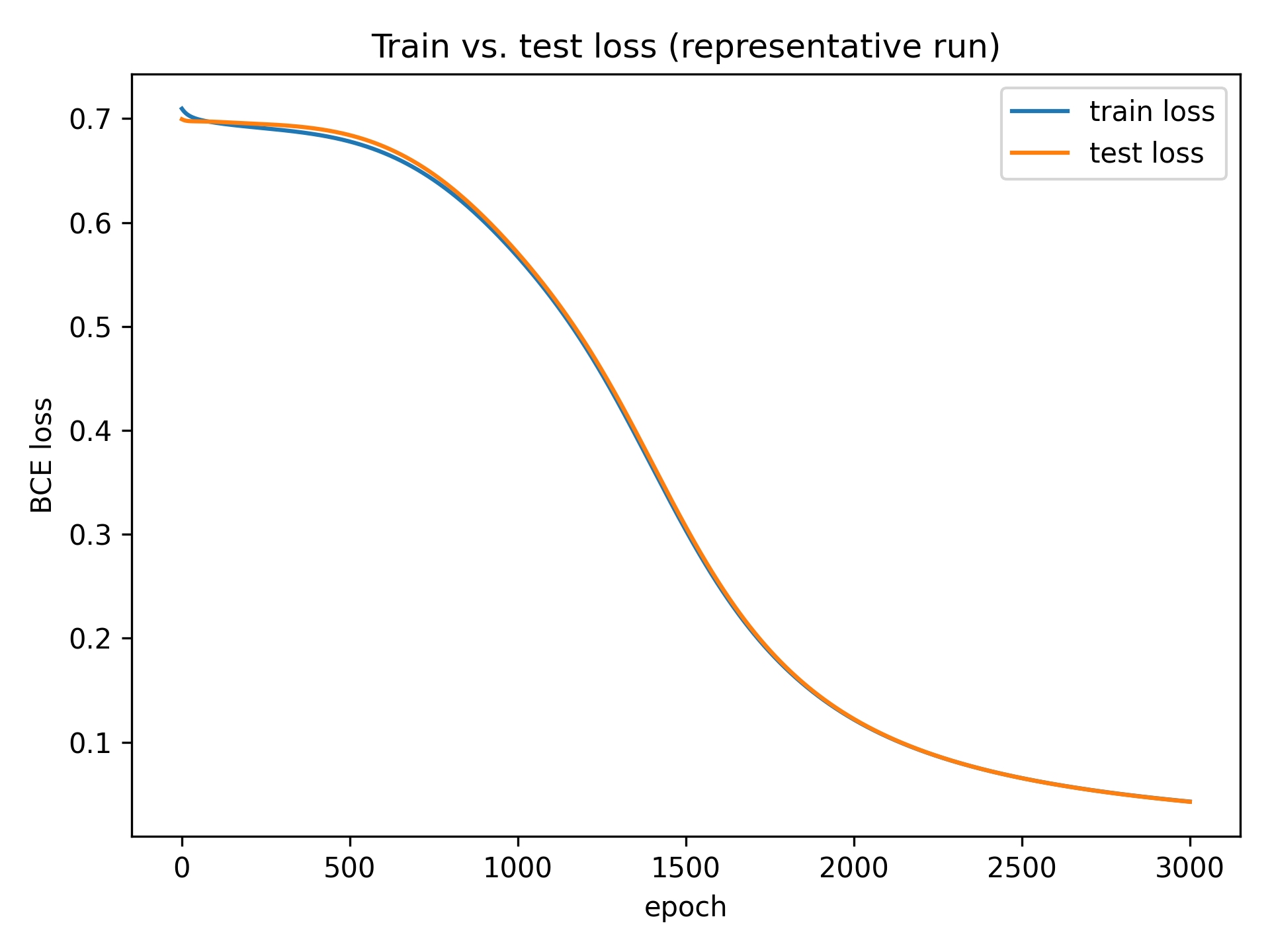}
  \caption{MLP ($h=4$): train vs.\ test loss}
  \label{fig:lc_mlp_loss}
\end{subfigure}
\hfill
\begin{subfigure}[t]{0.32\textwidth}
  \centering
  \includegraphics[width=\linewidth]{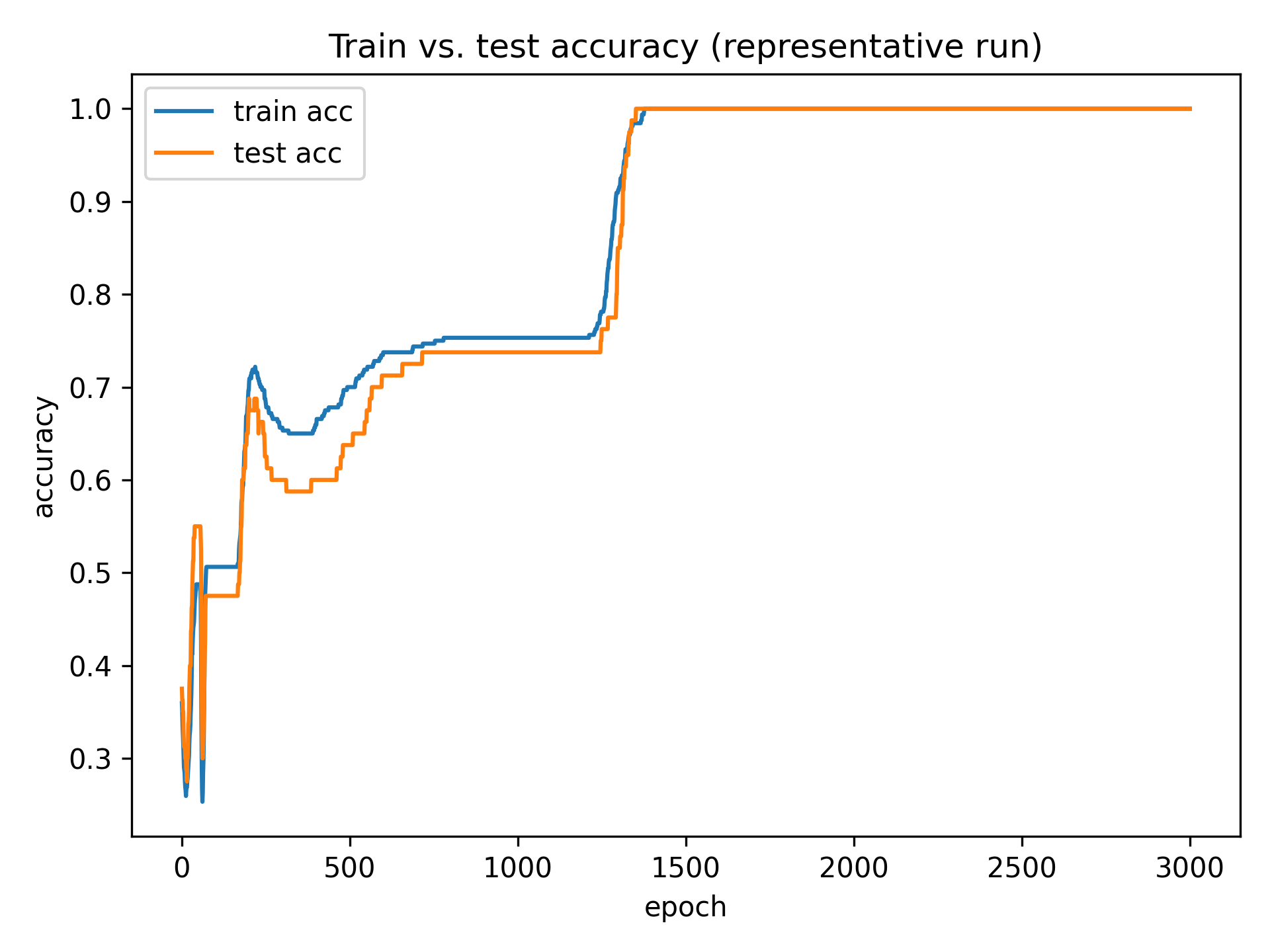}
  \caption{MLP ($h=4$): train vs.\ test accuracy}
  \label{fig:lc_mlp_acc}
\end{subfigure}
\hfill
\begin{subfigure}[t]{0.32\textwidth}
  \centering
  \includegraphics[width=\linewidth]
{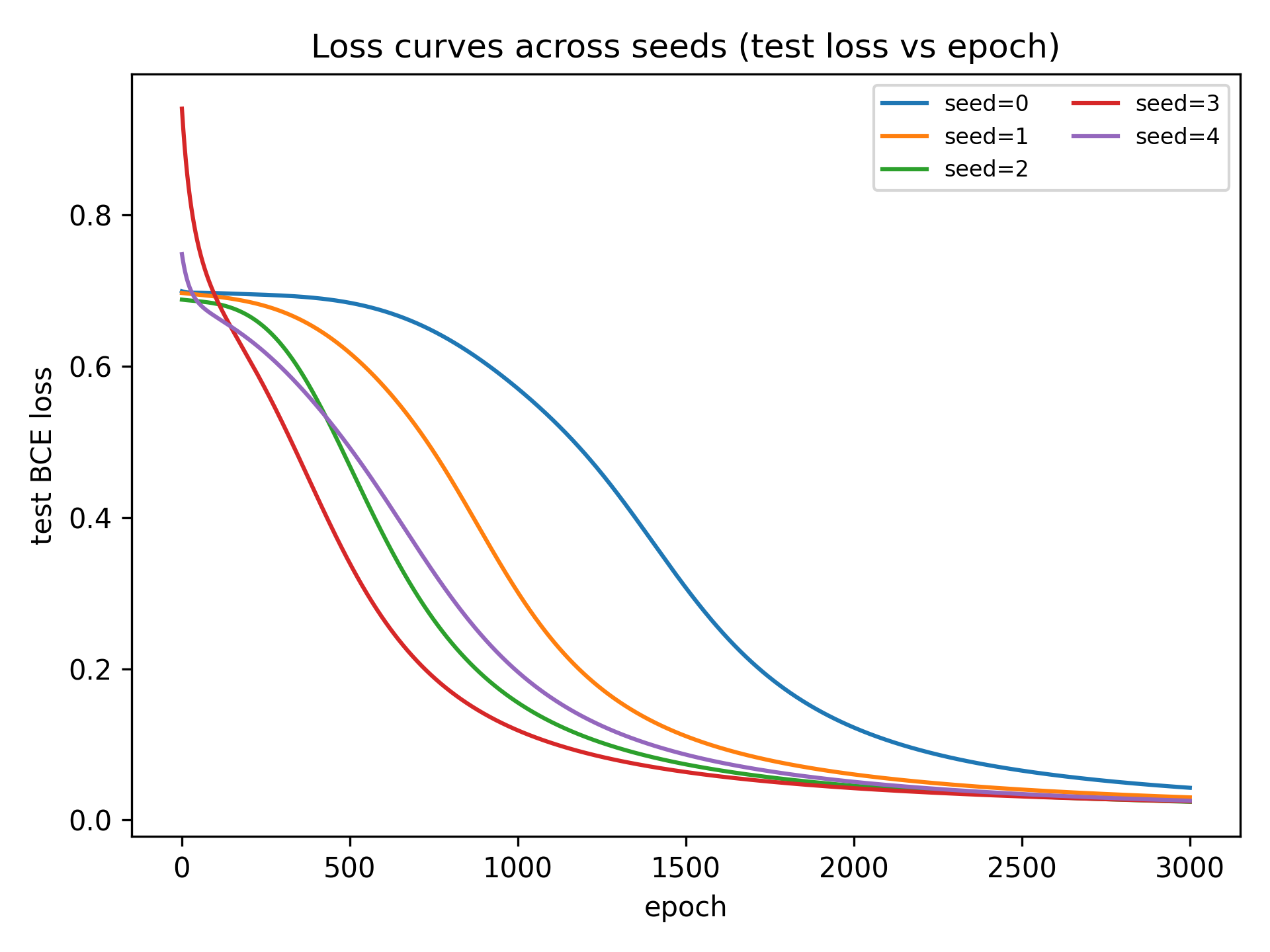}
  \caption{MLP ($h=4$): test loss across seeds}
  \label{fig:lc_mlp_seeds}
\end{subfigure}

\vspace{0.4em}

% ================== ROW 2 ==================
\begin{subfigure}[t]{0.32\textwidth}
  \centering
  \includegraphics[width=\linewidth]{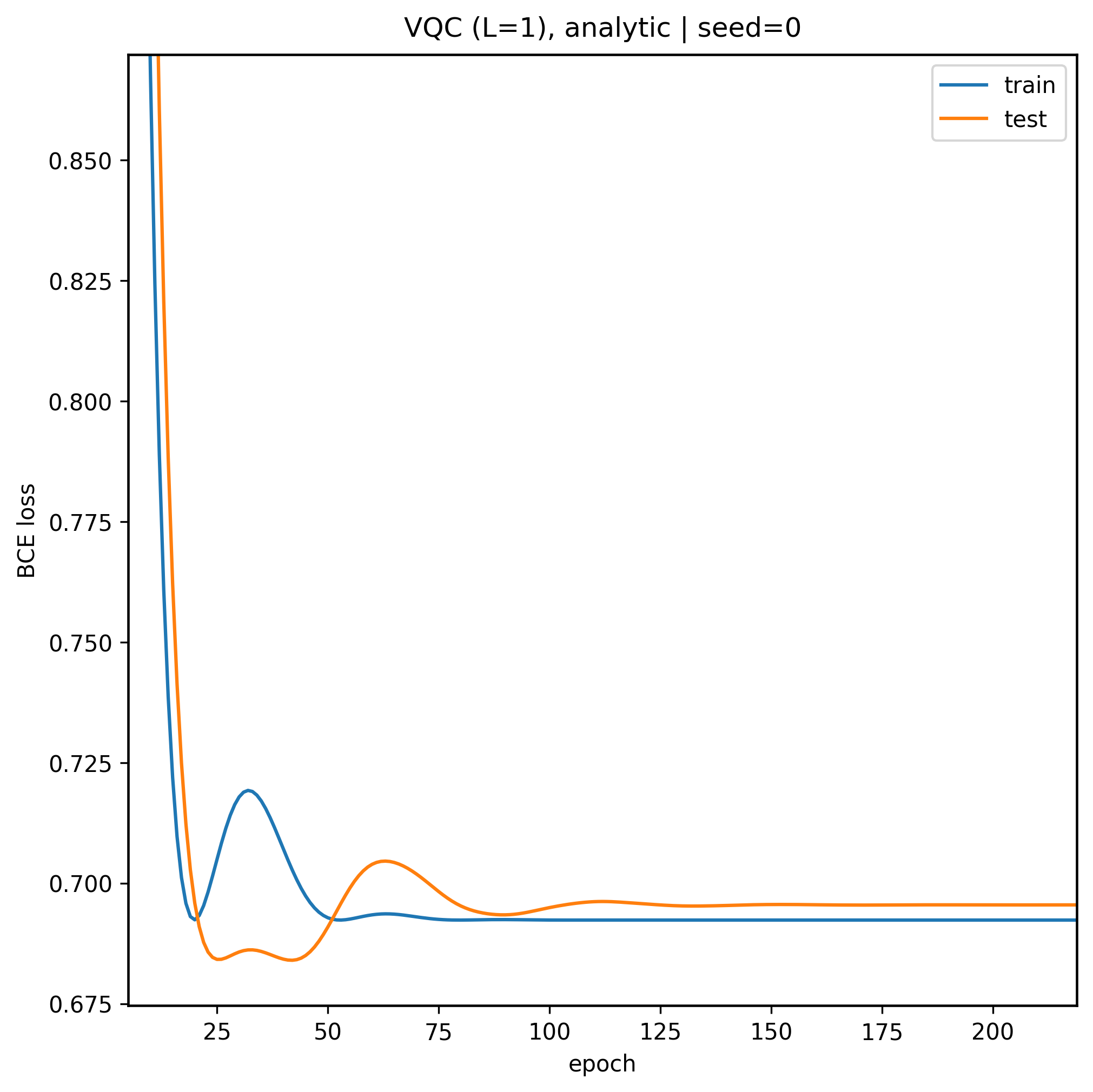}
  \caption{VQC ($L=1$): analytic}
  \label{fig:lc_vqc_L1_analytic}
\end{subfigure}
\hfill
\begin{subfigure}[t]{0.32\textwidth}
  \centering
  \includegraphics[width=\linewidth]{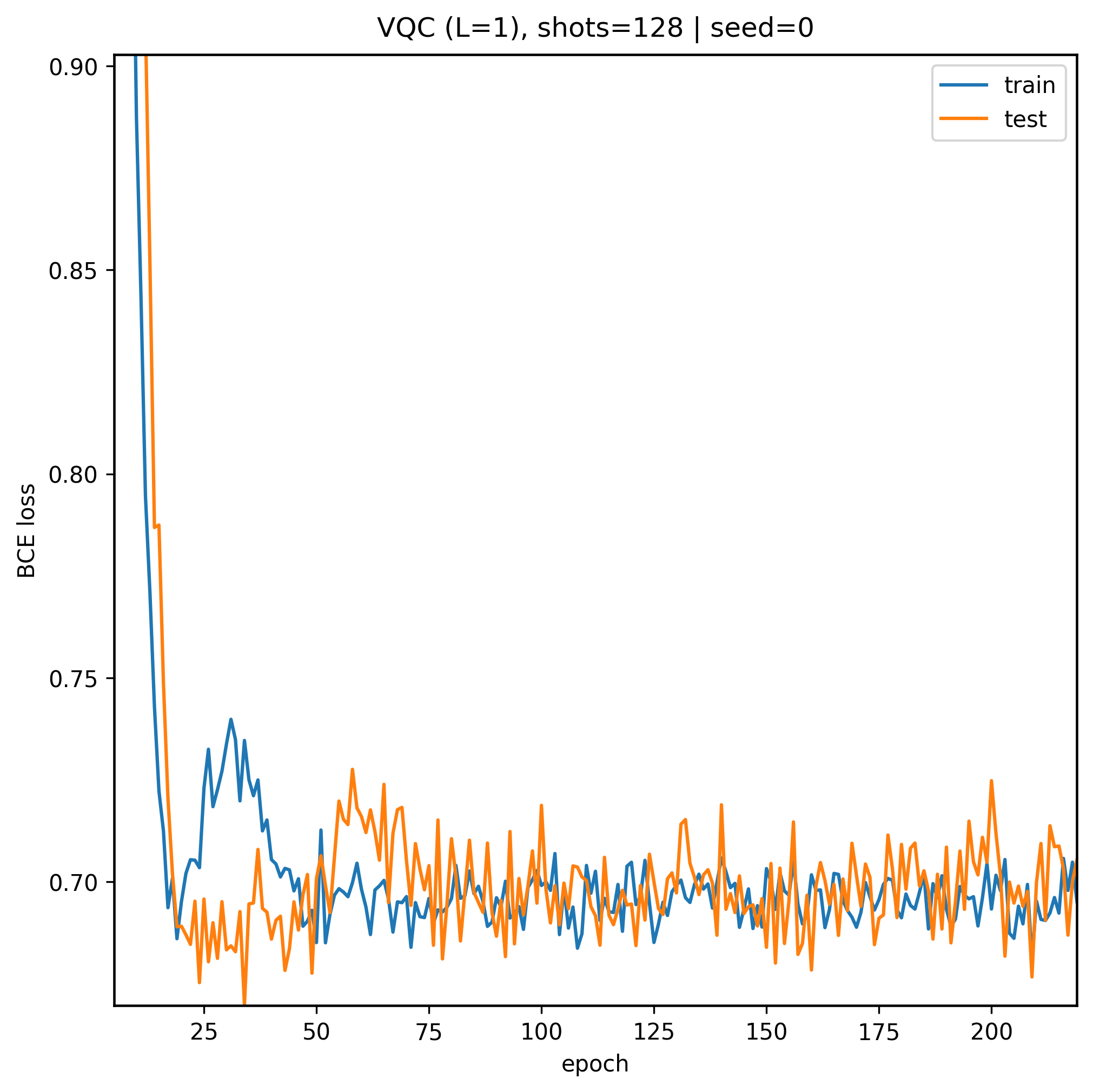}
  \caption{VQC ($L=1$): finite shots}
  \label{fig:lc_vqc_L1_shots}
\end{subfigure}
\hfill
\begin{subfigure}[t]{0.32\textwidth}
  \centering
  \includegraphics[width=\linewidth]
  {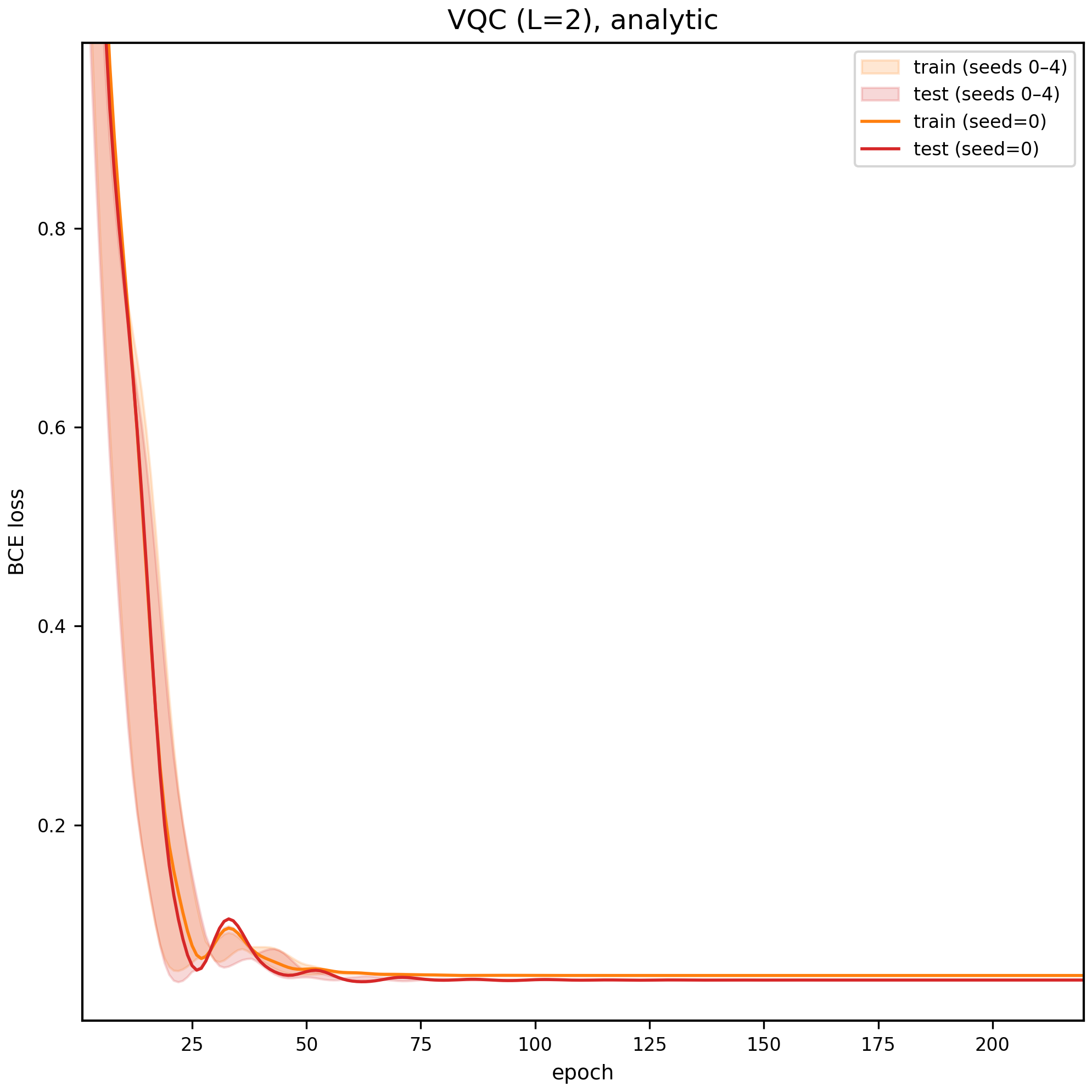}
  \caption{VQC ($L=2$): analytic}
  \label{fig:lc_vqc_seeds}
\end{subfigure}
\caption{Learning curves for the MLP ($h=4$) and the variational quantum
classifier (VQC), including training and test loss, accuracy, and representative
runs across random initializations.}
\label{fig:learning_behavior}
\end{figure}

For the MLP, panel~(a) shows the evolution of training and test loss as a function
of training epochs.
The training loss decreases continuously over the course of training and reaches
low values by the final epochs, while the test loss follows a similar trajectory
and remains close to the training loss throughout training.
Panel~(b) reports the corresponding training and test accuracy curves, which
increase progressively with training epochs and attain similar values at later
stages of training.
Panel~(c) shows test loss curves obtained from multiple random parameter
initializations, illustrating differences between trajectories during the early
training phase and variations in the rate of loss decrease, followed by comparable
loss values at later epochs.

For the VQC, panels~(d)--(f) report learning curves for different circuit depths
and measurement regimes.
Panel~(d) shows training and test loss curves for circuit depth $L=1$ in the
analytic setting, where both curves decrease over training epochs and approach
similar loss levels.
Panel~(e) presents the corresponding training and test loss curves for circuit
depth $L=1$ obtained using finite-shot estimation with $128$ shots, displaying an
overall decreasing trend with visible epoch-to-epoch variability.
Panel~(f) shows test loss curves for circuit depth $L=2$ in the analytic setting,
reported across multiple random parameter initializations, where trajectories
differ at early epochs and become closer at later stages of training.

\subsection{Robustness Analysis}
\label{subsec:robustness_all}

In this section, we present experimental results of the XOR task for the Linear, MLP(h=4), VQC(L=1), and VQC(L=2) models, under variations of noise level, dataset size, number of measurement shots, and random initialization.

\begin{figure}[!t]
\centering
\captionsetup[subfigure]{font=footnotesize,skip=2pt}

\begin{subfigure}[t]{0.32\textwidth}
  \centering
  \includegraphics[width=\linewidth]{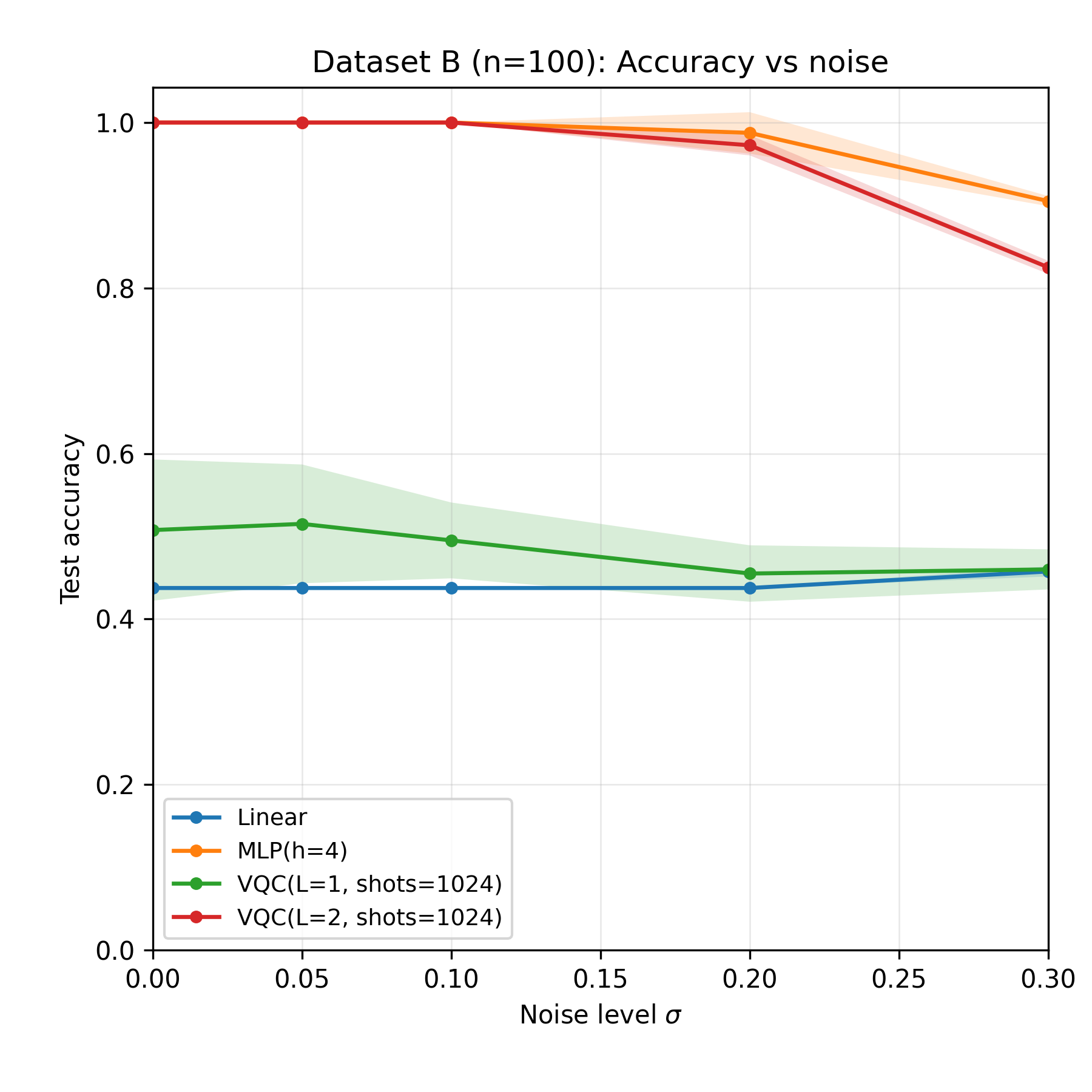}
  \caption{Accuracy vs.\ noise $\sigma$}
  \label{fig:rob_all_noise}
\end{subfigure}
\hfill
\begin{subfigure}[t]{0.32\textwidth}
  \centering
  \includegraphics[width=\linewidth]{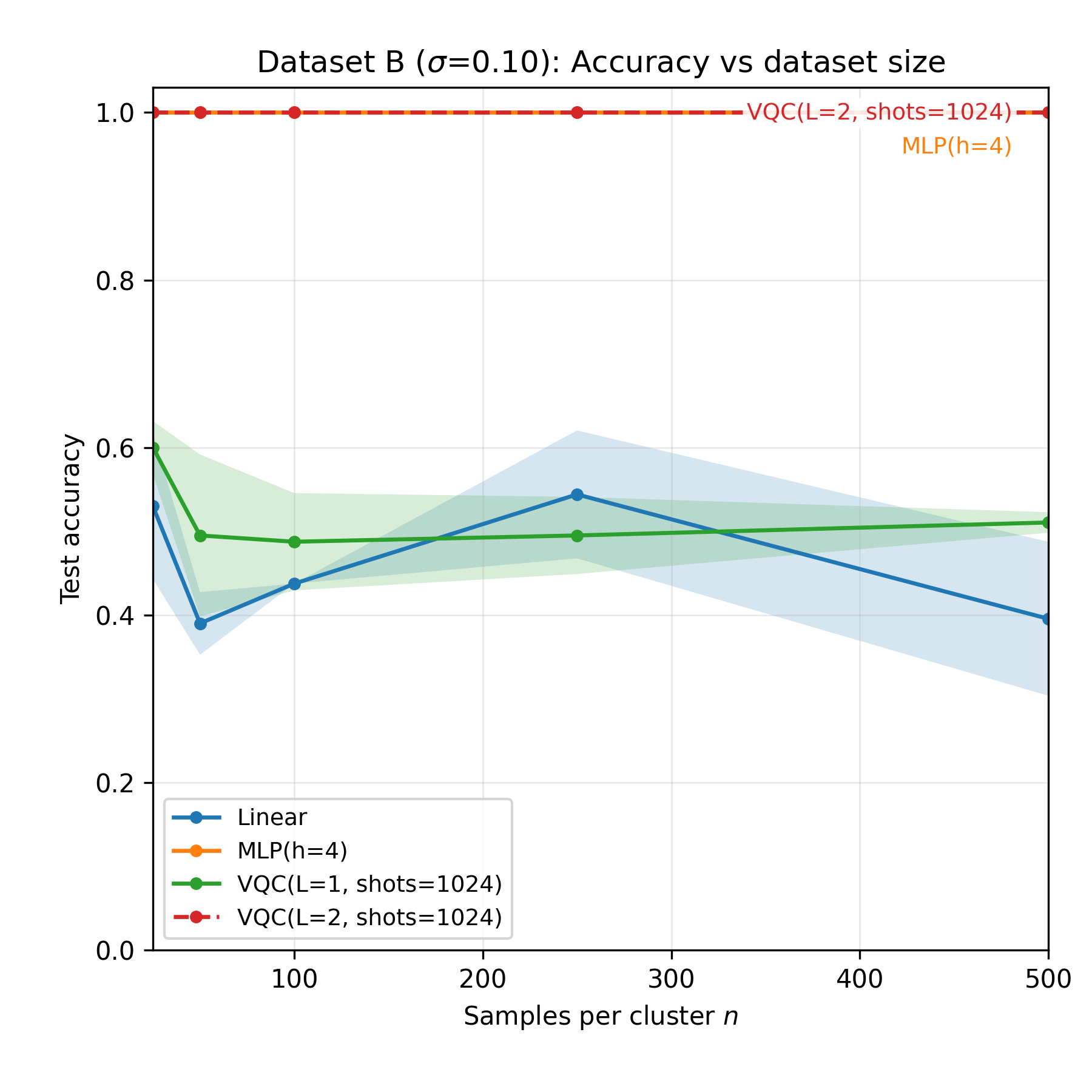}
  \caption{Accuracy vs.\ dataset size}
  \label{fig:rob_all_size}
\end{subfigure}
\hfill
\begin{subfigure}[t]{0.32\textwidth}
  \centering
  \includegraphics[width=\linewidth]{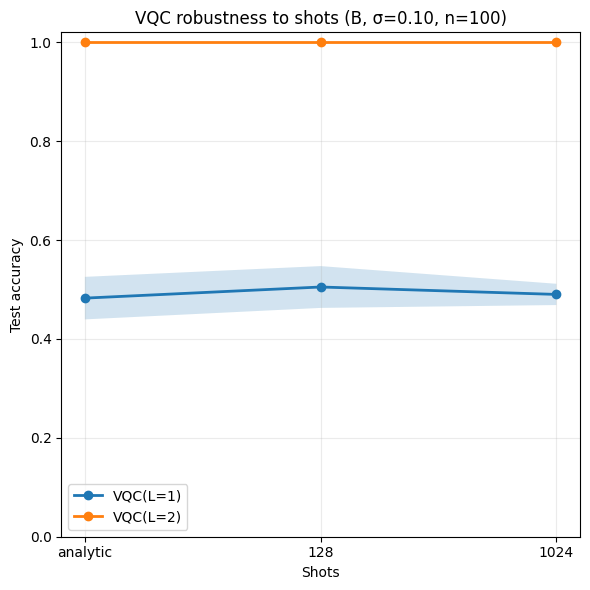}
  \caption{VQC accuracy vs.\ shots}
  \label{fig:rob_vqc_shots_only}
\end{subfigure}

\caption{Robustness of classical and quantum models.
Test accuracy as a function of data noise level, dataset size, and
measurement shot noise (VQC).}
\label{fig:robustness_combined}
\end{figure}

\begin{figure}[!htbp]
\centering
\begin{subfigure}[t]{0.495\textwidth}
  \centering
  \includegraphics[width=\linewidth]{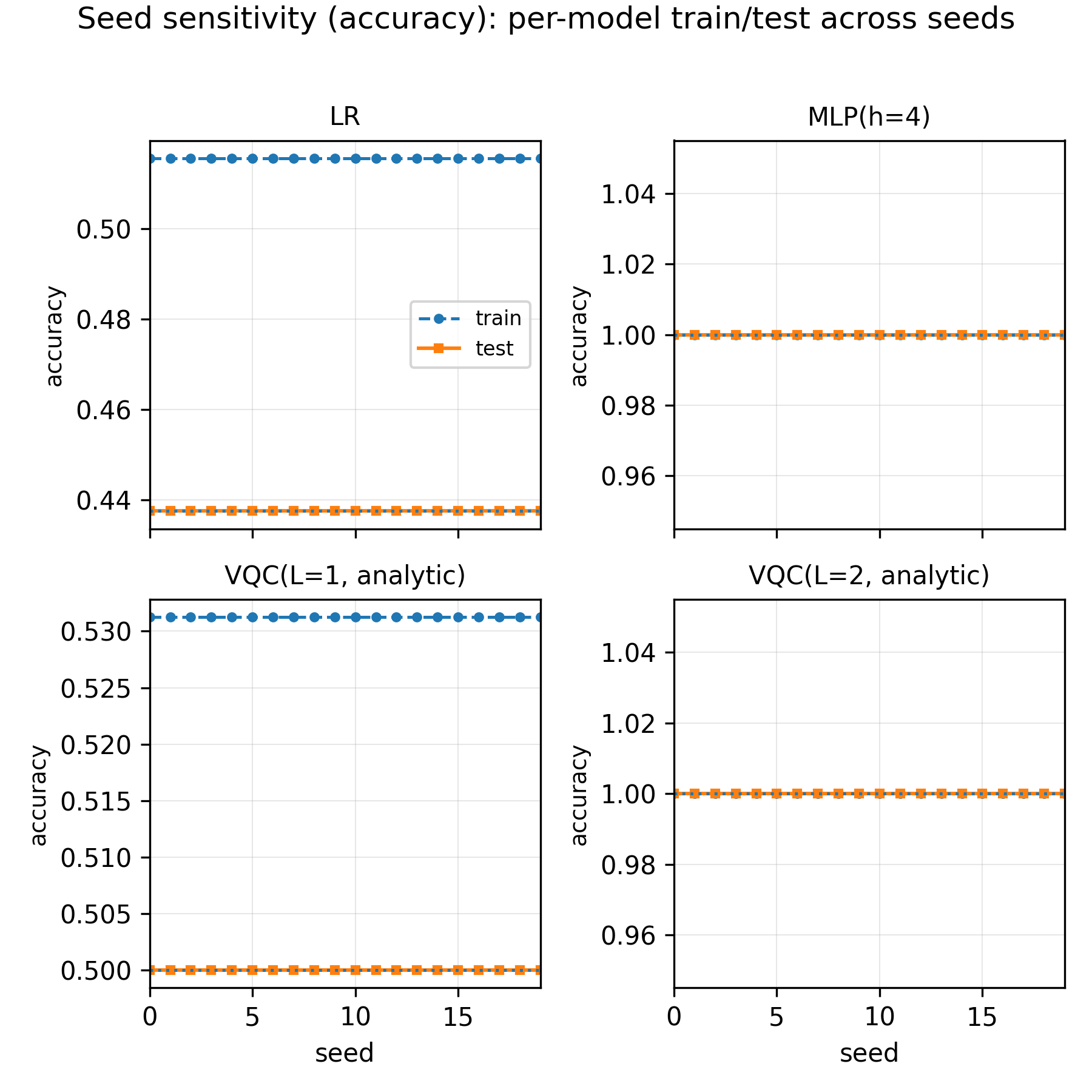}
  \caption{Seed sensitivity (accuracy)}
  \label{fig:seed_sens_acc_all}
\end{subfigure}
\hfill
\begin{subfigure}[t]{0.495\textwidth}
  \centering
  \includegraphics[width=\linewidth]{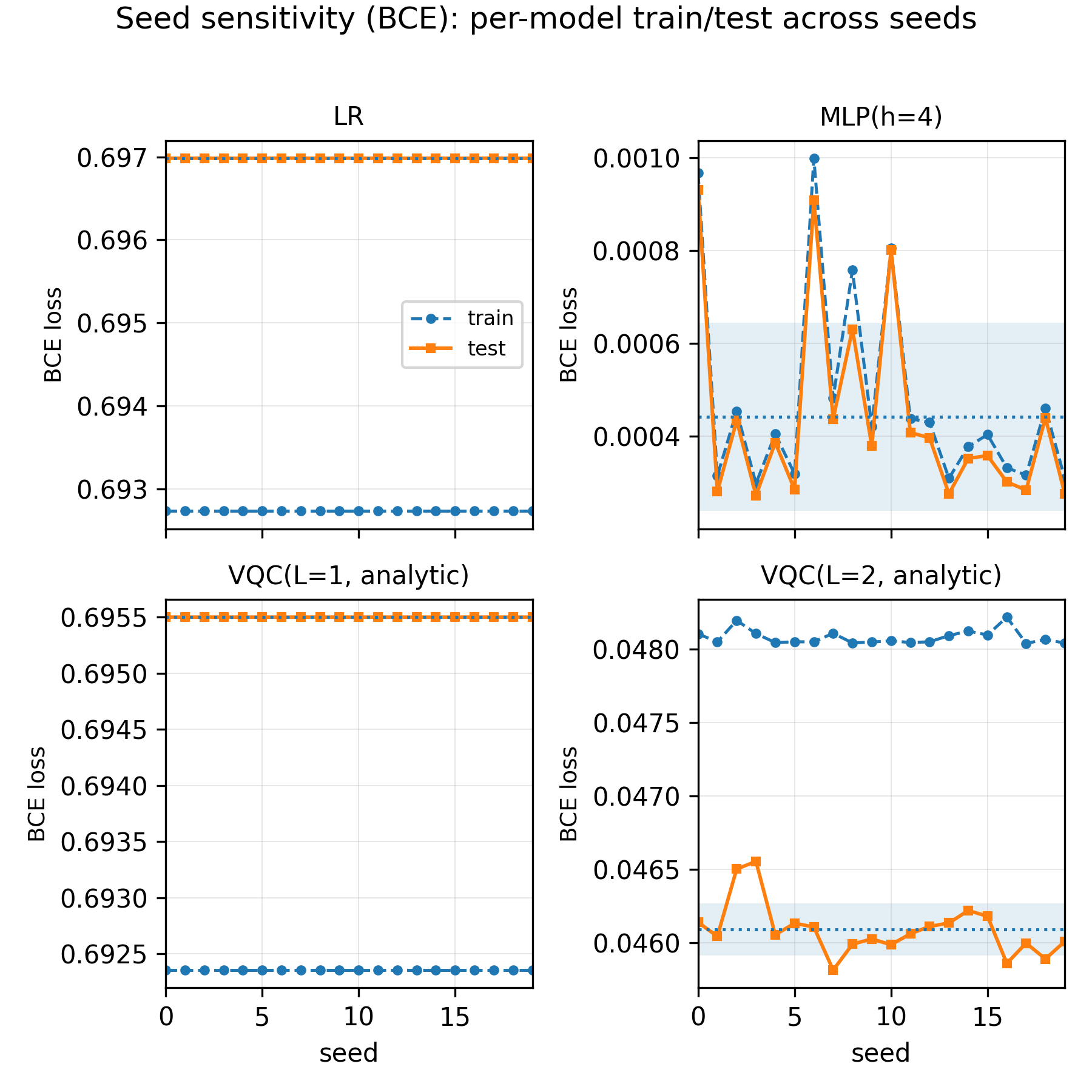}
  \caption{Seed sensitivity (BCE)}
  \label{fig:seed_sens_bce_all}
\end{subfigure}
\caption{Sensitivity to random initialization across models: accuracy and BCE evaluated over multiple training seeds.}

\label{fig:seed_sensitivity_all}
\end{figure}

Figure~\ref{fig:robustness_combined} (a) shows test accuracy as a function of the noise level $\sigma$. The linear model maintains low and nearly constant accuracy across the entire noise range. VQC(L=1) exhibits moderate accuracy values with a smooth decrease as noise increases. MLP(h=4) and VQC(L=2) coincide at low noise levels and reach maximal accuracy. As noise increases, both models show a gradual degradation in accuracy; at the highest noise level, the MLP(h=4) curve lies above the VQC(L=2) curve. Across the full range of $\sigma$, these two models achieve the highest accuracy among all models considered.

Figure~\ref{fig:robustness_combined} (b) shows test accuracy as a function of dataset size. For the linear model and VQC(L=1), the curves lie substantially lower and change only weakly as the number of samples per claster increases. For MLP(h=4) and VQC(L=2), accuracy coincides and equals 1.0 at all shown dataset sizes. 

Figure~\ref{fig:robustness_combined} (c) reports VQC accuracy as a function of the number of measurement shots. For VQC(L=2), the accuracy values are identical for the analytic regime and for finite-shot settings and coincide at all plotted points. For VQC(L=1), accuracy is lower and changes only slightly between the analytic, 128-shot, and 1024-shot regimes. The variability bands in the plot are staying narrow.

Figure~\ref{fig:seed_sensitivity_all} shows sensitivity to random initialization. In the accuracy–vs–seed plots, all models exhibit nearly horizontal curves. For MLP(h=4) and VQC(L=2), train and test accuracy coincide and remain equal to 1.0 across all runs. For the linear model and VQC(L=1), the accuracy values are lower but likewise stable across seeds, with no visible fluctuations.

In the BCE–vs–seed plots, the linear model and VQC(L=1) show nearly constant values across runs. MLP(h=4) and VQC(L=2) display a small spread of BCE values across seeds. At the same time, the BCE level of MLP(h=4) is consistently lower than that of VQC(L=2) over the entire plot, i.e., the classical variational model attains smaller BCE values while both models have equally maximal accuracy. Overall, in the seed-sensitivity and noise-robustness plots, MLP(h=4) and VQC(L=2) show similarly stable behavior and clearly outperform the simpler models.

\subsection{Summary Tables}
\label{subsec:summary_all}

Finally, we summarize model size, training cost, and predictive performance
from a dedicated benchmark run on Dataset~B
($\sigma=0.10$, $n_{\text{per cluster}}=100$).
Results are reported as mean $\pm$ std over five random initializations
(seeds $0$--$4$).

\begin{table}[!h]
\caption{Model performance on Dataset~B ($\sigma=0.10$, $n=100$).
Reported values are mean $\pm$ std over five model seeds ($0$--$4$).}
\label{tab:exp_performance}
\centering
\small
\setlength{\tabcolsep}{4pt}

\begin{tabularx}{\linewidth}{X c c c}
\hline
Model & Train acc. & Test acc. & Test loss (BCE) \\
\hline
LR
& 0.759 $\pm$ 0.000
& 0.713 $\pm$ 0.000
& 0.696 $\pm$ 0.000 \\

MLP ($h=4$)
& 1.000 $\pm$ 0.000
& 1.000 $\pm$ 0.000
& 0.016 $\pm$ 0.005 \\

VQC ($L=1$, analytic)
& 0.529 $\pm$ 0.004
& 0.508 $\pm$ 0.007
& 0.692 $\pm$ 0.000 \\

VQC ($L=1$, 1024 shots)
& 0.531 $\pm$ 0.018
& 0.525 $\pm$ 0.058
& 0.692 $\pm$ 0.003 \\

VQC ($L=1$, 128 shots)
& 0.496 $\pm$ 0.016
& 0.488 $\pm$ 0.056
& 0.696 $\pm$ 0.009 \\

VQC ($L=2$, analytic)
& 1.000 $\pm$ 0.000
& 1.000 $\pm$ 0.000
& 0.0185 $\pm$ 0.0000 \\

VQC ($L=2$, 1024 shots)
& 1.000 $\pm$ 0.000
& 1.000 $\pm$ 0.000
& 0.0187 $\pm$ 0.0006 \\

\hline
\end{tabularx}
\end{table}

\vspace{-1em}

\begin{table}[!htbp]
\caption{Model complexity and training cost on Dataset~B
($\sigma=0.10$, $n=100$).
Reported values are mean $\pm$ std over five model seeds.}
\label{tab:exp_cost}
\centering
\small
\setlength{\tabcolsep}{4pt}

\begin{tabularx}{\linewidth}{X c c}
\hline
Model & \#params & Train time (s) \\
\hline
LR
& 3
& 0.005 $\pm$ 0.000 \\

MLP ($h=4$)
& 17
& 0.053 $\pm$ 0.001 \\

VQC ($L=1$, analytic)
& 6
& 360.500 $\pm$ 4.500 \\

VQC ($L=1$, 1024 shots)
& 6
& 442.500 $\pm$ 3.700 \\

VQC ($L=1$, 128 shots)
& 6
& 410.880 $\pm$ 1.730 \\

VQC ($L=2$, analytic)
& 12
& 943.300 $\pm$ 7.300 \\

VQC ($L=2$, 1024 shots)
& 12
& 1776.100 $\pm$ 1503.800 \\

\hline
\end{tabularx}
\end{table}

The reported training times reflect wall-clock runtime in our software stack (optimized classical libraries for the MLP and quantum circuit simulation for the VQC) and are implementation-dependent. They should not be interpreted as a hardware-agnostic complexity comparison between the models.

\subsection{Topology and Expressivity Analysis}
\label{subsec:topology}

Beyond robustness and accuracy trends, we analyze how the topology of the
variational quantum classifier (VQC) influences representational capacity and
decision geometry. In particular, we focus on the role of circuit depth and its
effect on the learned decision function under identical data and training
conditions.

\subsubsection{Ansatz Depth: Expressivity vs.\ Complexity}
\label{subsubsec:ansatz_depth}

We study how the ansatz depth $L$ affects the expressive power of the VQC.
Increasing the number of layers introduces additional trainable rotations and
entangling operations, thereby enlarging the hypothesis class. For the considered
two-qubit architecture, the number of trainable parameters scales linearly as
$6L$.

Figure~\ref{fig:vqc_depth_db} shows representative learned decision surfaces for
depths $L=1$ and $L=2$ in the analytic regime. With a single layer, the circuit
produces only a limited variation of the output landscape and fails to form a
clear nonlinear separation structure. In contrast, the deeper circuit ($L=2$)
yields a sharply structured, nonlinear decision surface consistent with the XOR
pattern.

This qualitative difference matches the quantitative results reported earlier:
shallow circuits tend to underfit the XOR structure, whereas depth $L=2$ provides
sufficient expressivity for stable separation. At the same time, increased depth
implies higher parameter count and greater computational cost per optimization
step, so the gain in expressivity should be evaluated relative to task
complexity.

\begin{figure}[h]
\centering
\begin{subfigure}[t]{0.495\textwidth}
  \centering
  \includegraphics[width=\linewidth]{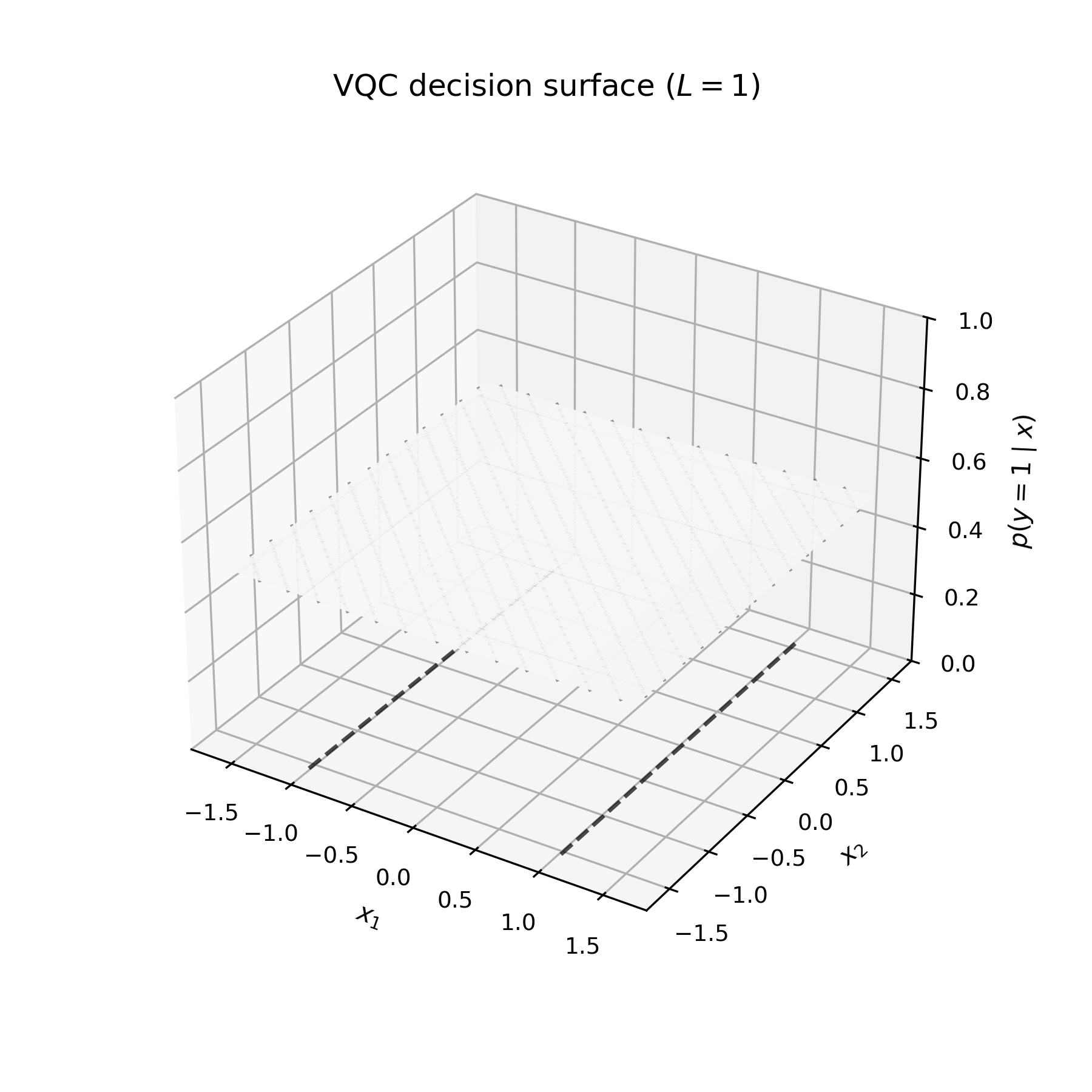}
  \caption{VQC decision surface ($L=1$)}
  \label{fig:db_vqc_L1_depth}
\end{subfigure}
\hfill
\begin{subfigure}[t]{0.495\textwidth}
  \centering
  \includegraphics[width=\linewidth]{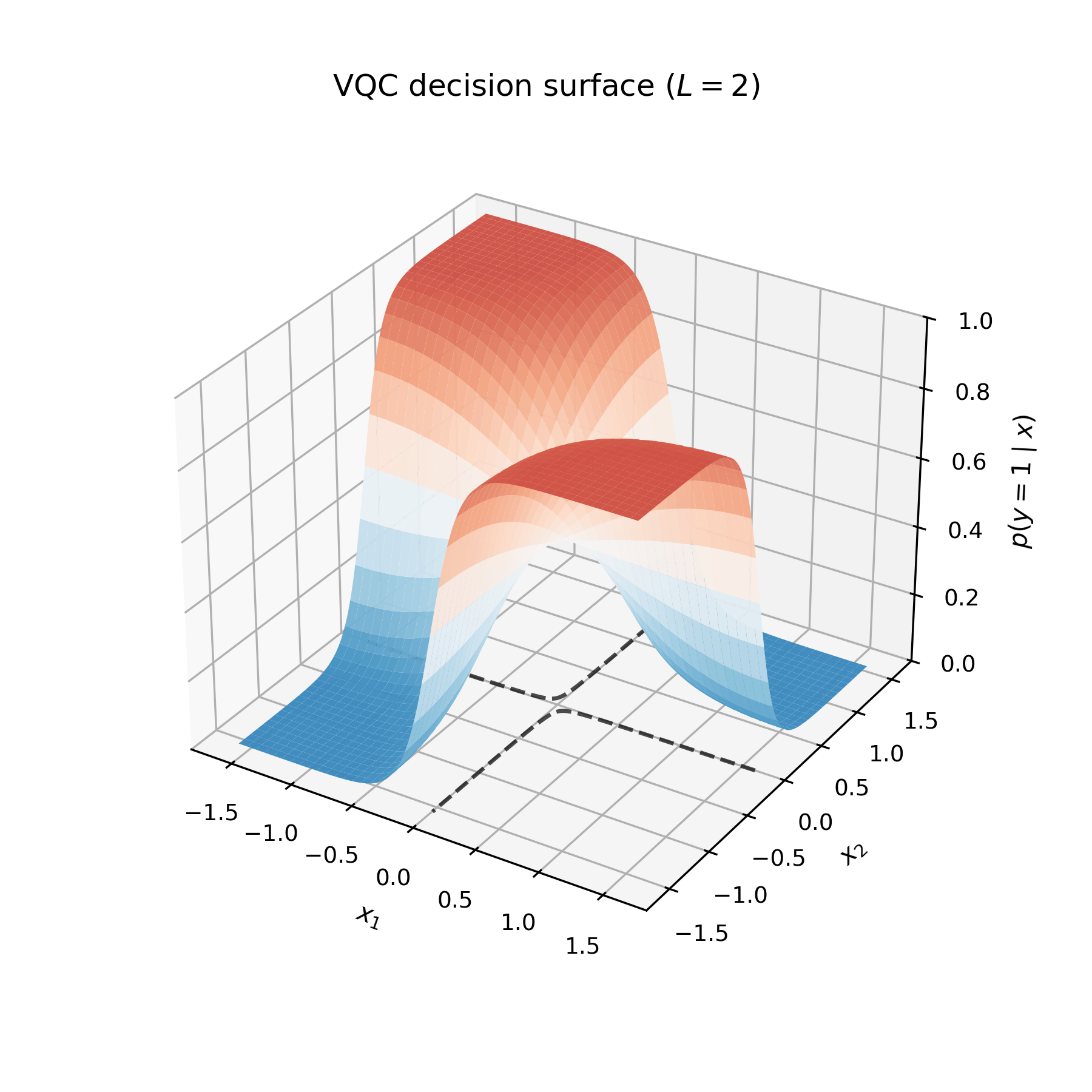}
  \caption{VQC decision surface ($L=2$)}
  \label{fig:db_vqc_L2_depth}
\end{subfigure}
\caption{Decision surfaces of the variational quantum classifier for circuit
depths $L=1$ and $L=2$ in the analytic regime. The color scale is consistent
with Fig.~\ref{fig:db_vqc_group_2x2}.}
\label{fig:vqc_depth_db}
\end{figure}

\subsubsection{Parameter Scaling with Depth}
\label{subsubsec:param_time}

The increase in circuit depth directly affects model size. In the considered
ansatz, each additional layer contributes one rotation block per qubit together
with an entangling operation, leading to a linear parameter scaling of $6L$.
Thus, moving from $L=1$ to $L=2$ doubles the number of trainable parameters and
expands the accessible function class.

Empirically, this increase in parameterization correlates with a transition from
near-linear or weakly nonlinear decision behavior to a fully nonlinear decision
surface capable of resolving the XOR structure. This supports the interpretation
that circuit depth acts as an effective expressivity control knob in small-scale
VQC models.

A complementary qualitative analysis based on low-dimensional loss landscape
slices is provided in Appendix~\ref{app:loss_landscape}.
These visualizations indicate broadly similar landscape structure across circuit
depths, while reflecting differences in the attained loss values.

\subsection{Simulation vs.\ Real Quantum Hardware}
\label{subsec:hardware}

To assess the practical impact of device-level noise beyond idealized simulation,
we evaluate the trained VQC on a real IBM Quantum backend~\cite{ibm_quantum,qiskit} and compare its
behavior against a matched simulator baseline.
All hardware experiments are performed in an \emph{inference-only} setting:
the VQC parameters are trained offline in simulation and then kept fixed while
executing the same circuit on hardware.
This isolates backend-related effects (gate and readout errors, decoherence~\cite{nielsen2010,preskill2018nisq},
and compilation-induced distortions) from training instability.

We focus on a controlled XOR setting (Dataset~A, clean XOR) and measure the
continuous prediction score
$f(x)=2p(y=1\mid x)-1$ on a dense evaluation grid (25$\times$25 points, 1024 shots per point),
using the same encoding and ansatz depth ($L=2$) as in the simulator experiments.
Although clean XOR can remain perfectly separable under moderate perturbations,
function-level comparisons reveal systematic deviations on hardware even when
accuracy is unchanged.

\begin{figure}[!htbp]
\centering
\captionsetup[subfigure]{font=footnotesize,skip=2pt}

\begin{subfigure}[t]{0.48\textwidth}
  \centering
  \includegraphics[width=\linewidth]{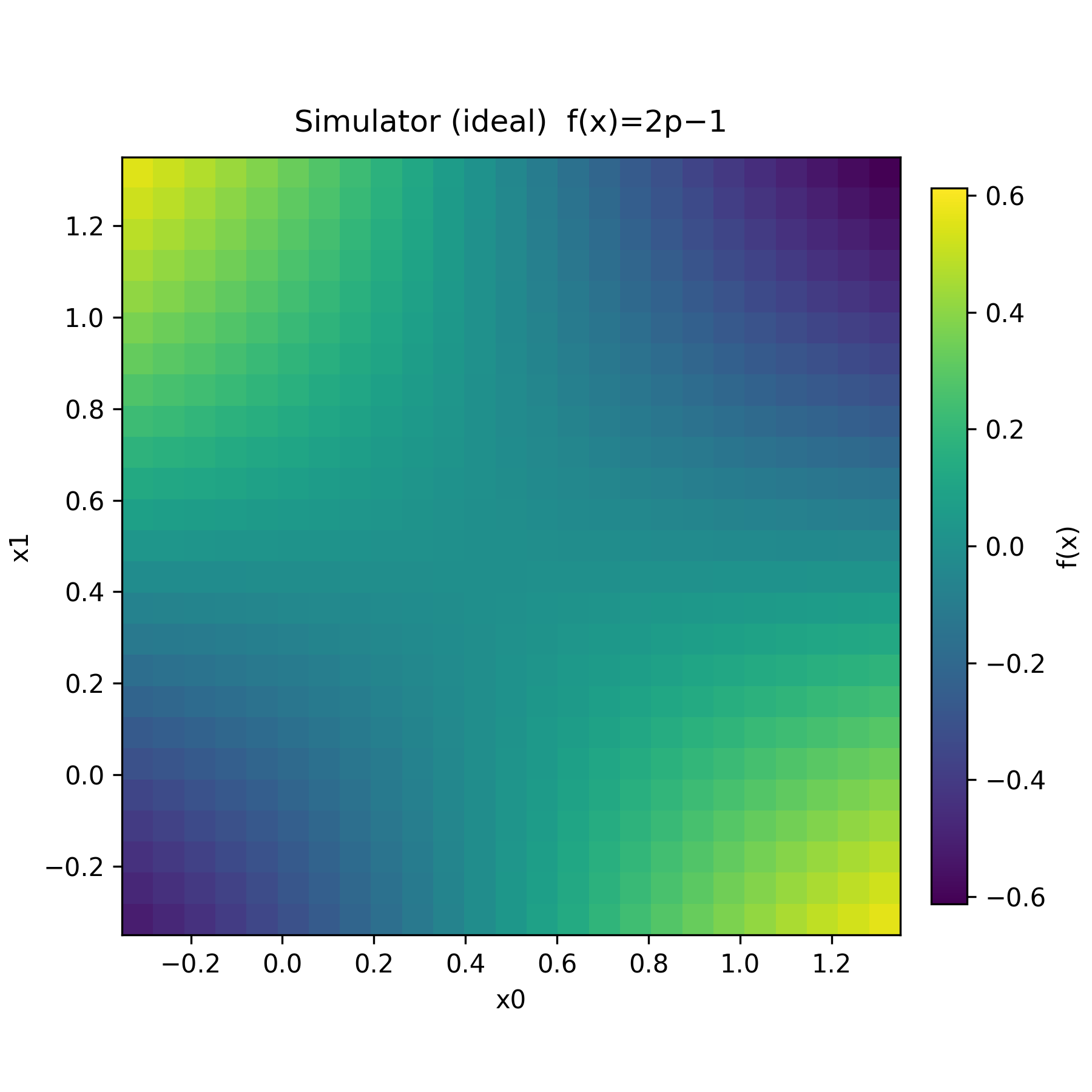}
  \caption{Simulator (ideal)}
  \label{fig:hw_f_sim}
\end{subfigure}
\hfill
\begin{subfigure}[t]{0.48\textwidth}
  \centering
  \includegraphics[width=\linewidth]{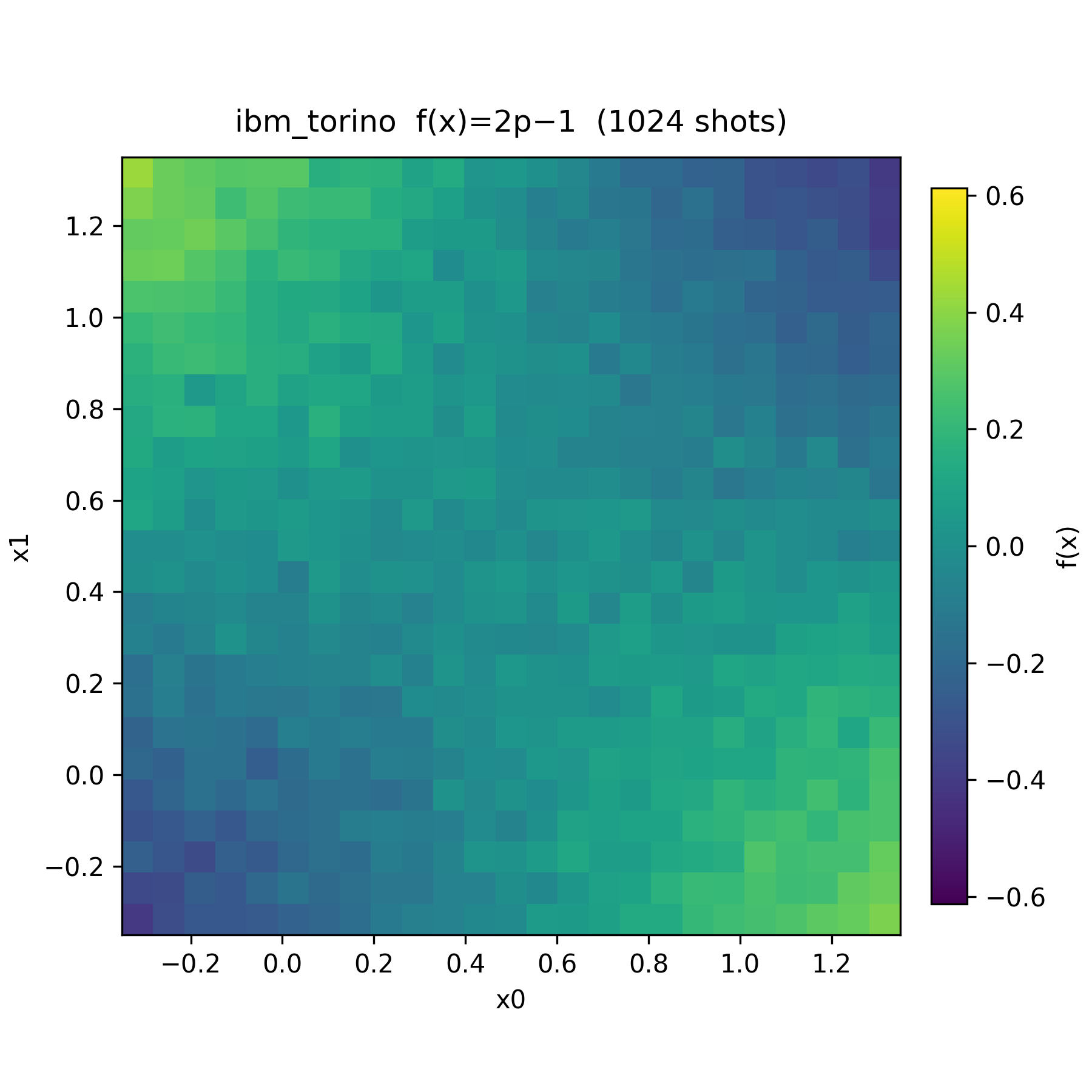}
  \caption{IBM hardware (1024 shots)}
  \label{fig:hw_f_hw}
\end{subfigure}

\caption{Decision function comparison for a fixed trained VQC ($L=2$) on clean XOR.
The color encodes $f(x)=2p(y=1\mid x)-1$ evaluated on a dense grid.
Hardware execution preserves the global XOR-like structure but introduces visible
local fluctuations and reduced smoothness compared to the ideal simulator.}
\label{fig:sim_vs_hw_fx}
\end{figure}

Figure~\ref{fig:sim_vs_hw_fx} shows that the global decision geometry learned in simulation
largely transfers to hardware: the sign structure of $f(x)$ remains consistent with
the expected XOR partitioning.
However, the hardware surface exhibits increased granularity and local irregularities,
consistent with device-level noise sources that are not captured by analytic expectation values.

To quantify these deviations, we analyze pointwise distortions and summary statistics.
In particular, we compute the absolute difference map $|f_{\mathrm{hw}}(x)-f_{\mathrm{sim}}(x)|$
over the full evaluation grid and compare shot noise against hardware noise via a
controlled metric-based experiment.

\begin{figure}[t]
\centering
\captionsetup[subfigure]{font=footnotesize,skip=2pt}

\begin{subfigure}[t]{0.48\textwidth}
  \centering
  \includegraphics[width=\linewidth]{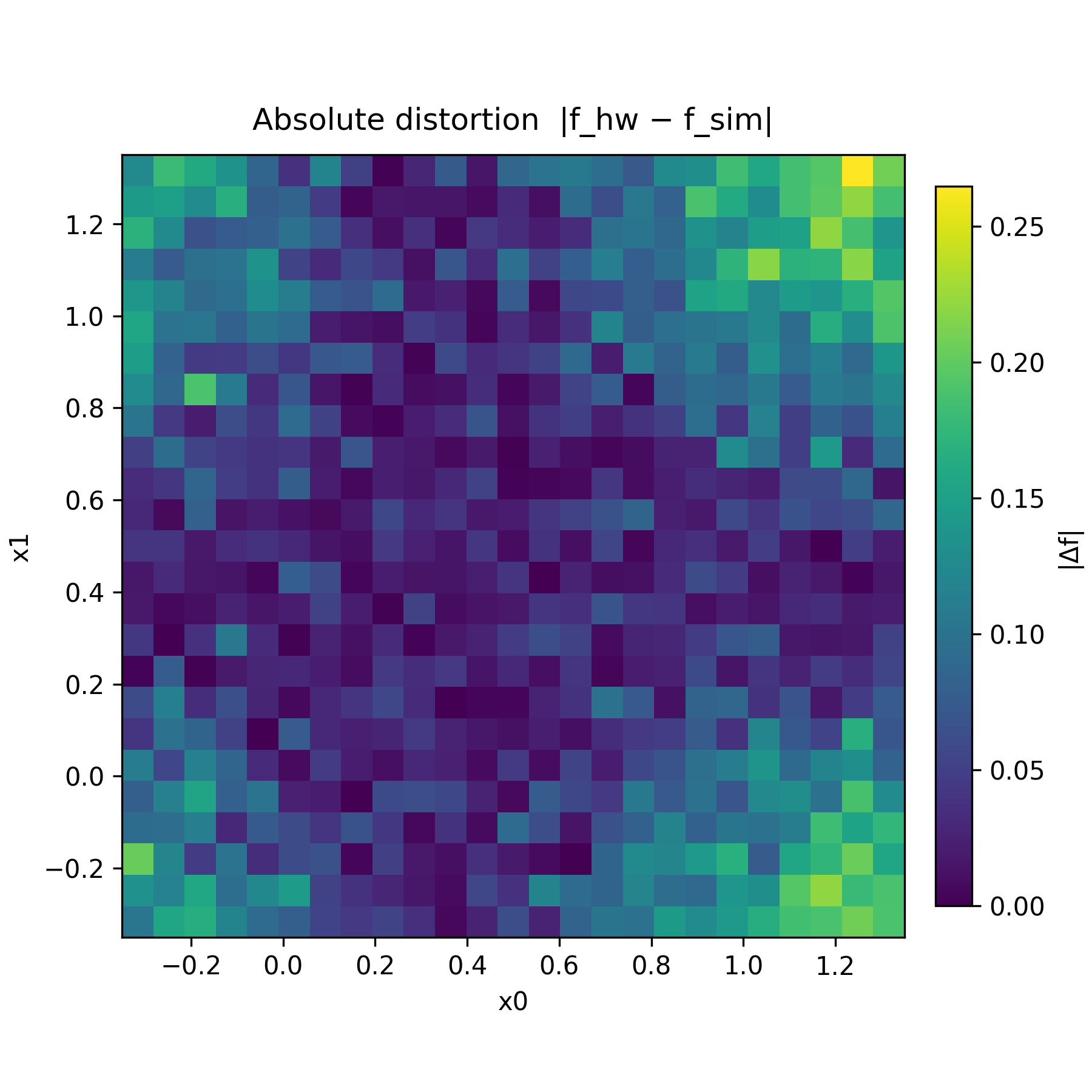}
  \caption{Pointwise distortion $|f_{\mathrm{hw}}-f_{\mathrm{sim}}|$}
  \label{fig:hw_absdiff}
\end{subfigure}
\hfill
\begin{subfigure}[t]{0.48\textwidth}
  \centering
  \includegraphics[width=\linewidth]{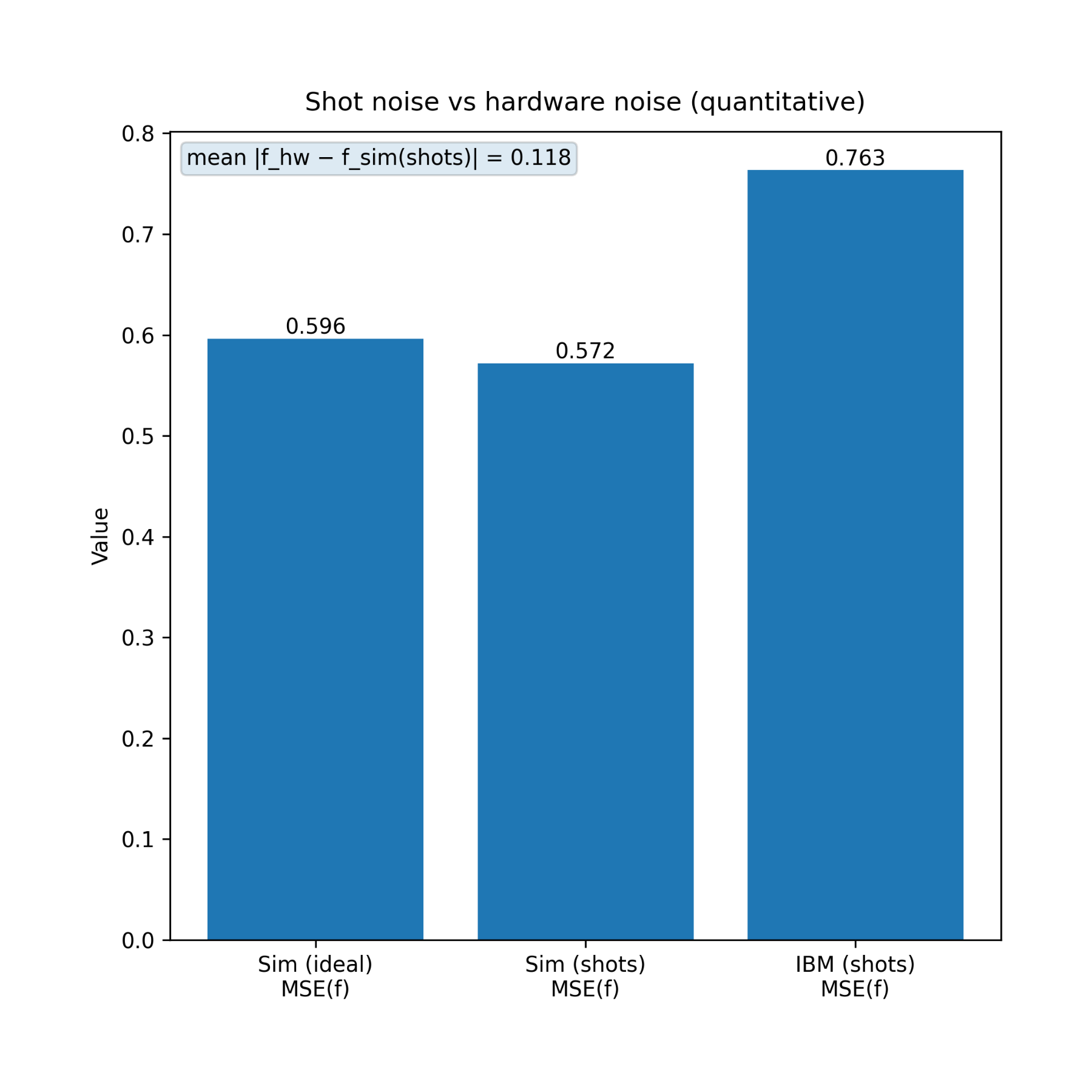}
  \caption{Shot noise vs.\ hardware noise (summary)}
  \label{fig:hw_mse_bars}
\end{subfigure}

\caption{Quantifying hardware-induced deviations for a fixed trained VQC.
Left: absolute distortion map over the evaluation grid, revealing structured,
spatially varying deviations (largest near the outer regions of the grid).
Right: controlled comparison showing that finite-shot simulation remains close to
the ideal baseline, whereas hardware execution increases the deviation magnitude.
In our run, the mean absolute deviation between hardware and finite-shot simulation is
$\approx 0.118$, while the MSE summary indicates a clear gap between shot noise and device noise.}
\label{fig:hw_quant}
\end{figure}

Figure~\ref{fig:hw_quant} (left) demonstrates that the mismatch between hardware and ideal simulation
is not purely uniform: the distortion varies across the input space and tends to increase
toward the edges of the evaluation region.
This is consistent with structured, backend-dependent errors (e.g., coherent gate errors,
readout bias, or compilation artifacts) rather than purely i.i.d.\ sampling noise.

Figure~\ref{fig:hw_quant} (right) provides a compact quantitative decomposition.
While finite-shot simulation (1024 shots) remains close to the ideal simulator baseline,
hardware execution produces a noticeably larger deviation, despite preserving perfect accuracy
on clean XOR in this setting.
This highlights an important practical point: for easy separable problems, accuracy alone can
mask substantial function-level distortions, which become visible when analyzing the continuous
decision score $f(x)$.

\paragraph{Additional diagnostics.}
A complementary distributional view of $f(x)$ (histogram comparison between simulator and hardware)
is provided in Appendix~\ref{app:hardware_additional}.

\section{Discussion}
\label{sec:discussion}

This study set out to answer a simple question: when we put classical and quantum
models side by side on the XOR problem, what actually makes the difference?
Although XOR is minimal, it is structurally nontrivial, and this makes it a clean
testbed for isolating architectural effects without confounding factors.

\subsection{Architectural Expressivity and Depth}

The clearest message from the experiments is that success on XOR is governed by
architectural expressivity. Whether a model is classical or quantum is secondary.
What matters is whether the architecture can represent a nonlinear decision
boundary of the required shape.

The linear classifier consistently remains near chance level (see
Section~\ref{subsec:robustness_all}), exactly as predicted by the theoretical
non-separability discussed in Section~\ref{sec:problem}. No amount of training
rescues it. In contrast, both the MLP and the VQC succeed---but only once their
structure is sufficiently expressive.

For the VQC, circuit depth is the decisive factor. With $L=1$, the learned
decision surfaces (Figures~\ref{fig:db_vqc_group_2x2} and
\ref{fig:vqc_depth_db}) fail to reproduce the XOR geometry even under analytic
evaluation. Increasing the depth to $L=2$ produces a qualitative shift: the
decision surface becomes fully nonlinear and aligns with the XOR structure. This
transition mirrors the classical case, where a minimum hidden-layer capacity is
required to solve XOR.

Importantly, this behavior is not explained by parameter count alone. The
VQC($L=1$) already contains trainable parameters, yet its circuit topology
restricts the function class it can represent. In this sense, topology dominates
raw parameter quantity.

\subsection{Classical vs.\ Quantum Performance}

A direct comparison between MLP($h=4$) and VQC($L=2$) shows that both models
achieve perfect test accuracy on the benchmark configuration (Dataset~B,
$\sigma=0.10$, $n=100$ per cluster; see Table~\ref{tab:exp_performance}). Across
the tested noise levels and dataset sizes
(Figure~\ref{fig:robustness_combined}), their maximal accuracies coincide.

However, accuracy alone does not tell the whole story.

First, the MLP consistently achieves lower binary cross-entropy (BCE) than
VQC($L=2$), even when both reach $1.0$ accuracy. This means that the MLP tends
to produce sharper probability estimates, while the VQC outputs remain smoother.
The difference is not visible in accuracy, but it reflects distinct predictive
behavior.

Second, the training cost differs by orders of magnitude in our implementation
(Table~\ref{tab:exp_cost}). Despite having fewer trainable parameters, the VQC
requires substantially longer training time than the MLP. We emphasize that this
comparison is implementation-dependent, yet it remains practically relevant:
under identical experimental conditions, the quantum model is markedly less
efficient to train.

Taken together, these results indicate that for low-dimensional tasks such as
XOR, deeper VQCs can match classical neural networks in accuracy but do not
provide a measurable advantage in loss-based performance or computational cost.

\subsection{Robustness and Noise Effects}

The robustness analysis (Section~\ref{subsec:robustness_all}) reveals similar
qualitative trends for MLP($h=4$) and VQC($L=2$). As Gaussian noise increases,
performance degrades smoothly rather than abruptly. Likewise, varying dataset
size does not alter the qualitative hierarchy of models. This suggests that
robustness is primarily a function of expressive capacity, not of model type.

Finite-shot evaluation introduces additional stochasticity in the VQC.
Nevertheless, the global XOR structure remains stable for $L=2$, and the
difference between analytic and finite-shot evaluation is quantitative rather
than structural (Figures~\ref{fig:robustness_combined} and
\ref{fig:db_vqc_group_2x2}).

Experiments on real IBM Quantum hardware further clarify this point. While the
global XOR pattern is preserved, the learned decision function exhibits
structured deviations relative to simulation, with a mean absolute difference of
approximately $0.118$ in a representative experiment. This discrepancy does not
necessarily reduce accuracy, but it highlights effects that are invisible to
discrete metrics alone.

\subsection{Limitations and Threats to Validity}

Several limitations should be kept in mind.

First, all experiments are conducted on synthetic, low-dimensional datasets.
The conclusions therefore apply strictly to controlled XOR benchmarks and may
not transfer directly to high-dimensional or real-world tasks.

Second, only a single VQC architecture with fixed encoding and two qubits is
studied. Alternative ansätze, encodings, or larger systems could alter the
observed trade-offs.

Third, optimizer choice and hyperparameters are fixed across experiments for
fairness, but different configurations may affect convergence speed or final
loss values.

Finally, hardware experiments are limited to inference on current NISQ devices.
Training directly on hardware, or scaling to deeper circuits and more qubits,
may introduce additional phenomena not captured here.

\subsection{Practical Implications}

From a practical standpoint, classical neural networks remain the more efficient
choice for simple nonlinear classification tasks. They achieve the same accuracy
as expressive VQCs while training faster and producing slightly better-calibrated
outputs. In this setting, the quantum model does not offer a performance
advantage.

At the same time, the VQC serves as a controlled laboratory for studying how
circuit depth, noise, and optimization interact. Even on a minimal task such as
XOR, these factors produce measurable structural effects that are not always
visible in accuracy metrics.

\subsection{Future Work}

Future work should move toward controlled scaling of task complexity, where the
difficulty increases gradually and classical models begin to encounter genuine
representational limits. Such settings are better suited for identifying regimes
in which quantum models may scale differently.

A deeper investigation of the VQC parameter landscape is also warranted,
particularly the relationship between circuit depth, loss geometry, and
optimization stability.

Finally, systematically studying the gap between simulation and real hardware as
circuit depth and qubit count grow remains essential. Hybrid architectures, in
which quantum circuits act as expressive feature maps within classical pipelines,
represent a promising direction for bridging theoretical potential and practical
utility.

\section{Conclusion}
\label{sec:conclusion}

In this work, we investigated the behavior of classical and quantum variational
classifiers on the XOR problem, focusing on expressivity, robustness, and
computational cost. The results demonstrate that the ability to solve XOR is
determined by architectural expressivity rather than by the use of quantum
computing per se.

\vspace{1em}
\noindent
When sufficiently expressive, variational quantum classifiers can match the
classification accuracy of classical neural networks and exhibit comparable
robustness trends with respect to data noise, dataset size, and random
initialization. However, they require substantially longer training times and do
not outperform classical models in terms of loss-based metrics. These findings
suggest that, for simple nonlinear tasks, quantum variational models do not yet
offer practical advantages over well-established classical approaches.

\vspace{1em}
\noindent
Experiments on real quantum hardware further show that, even when classification
accuracy is preserved, hardware execution introduces systematic, structured
deviations at the level of the continuous decision function that are not captured
by accuracy alone. This highlights the importance of function-level analysis when
assessing the practical impact of device-level noise.

\vspace{1em}

\noindent
Overall, this study clarifies the role of circuit depth, architectural design,
and noise effects in variational quantum learning and provides a rigorous baseline
for future investigations aimed at identifying problem regimes where quantum
models may exhibit genuine benefits.

\newpage
\bibliographystyle{splncs04}
\bibliography{references}

\begin{thebibliography}{10}
\providecommand{\url}[1]{\texttt{#1}}
\providecommand{\urlprefix}{URL }
\providecommand{\doi}[1]{https://doi.org/#1}

\bibitem{cerezo2021costdependent}
Cerezo, M., Sone, A., Volkoff, T., Cincio, L., Coles, P.J.: Cost function dependent barren plateaus in shallow parametrized quantum circuits. Nature Communications  \textbf{12}, ~1791 (2021). \doi{10.1038/s41467-021-21728-w}

\bibitem{havlicek2019quantumkernel}
Havl{\'\i}{\v c}ek, V., C{\'o}rcoles, A.D., Temme, K., Harrow, A.W., Kandala, A., Chow, J.M., Gambetta, J.M.: Supervised learning with quantum-enhanced feature spaces. Nature  \textbf{567},  209--212 (2019). \doi{10.1038/s41586-019-0980-2}

\bibitem{ibm_quantum}
{IBM}: Ibm quantum. \url{https://quantum.ibm.com/} (2026), accessed: 2026-02-09

\bibitem{larocca2025barren}
Larocca, M., Thanasilp, S., Wang, S., Sharma, K., Biamonte, J., Coles, P.J., Cincio, L., McClean, J.R., Holmes, Z., Cerezo, M.: Barren plateaus in variational quantum computing. Nature Reviews Physics  \textbf{7},  174--189 (2025). \doi{10.1038/s42254-025-00813-9}

\bibitem{mcclean2016theory}
McClean, J.R., Romero, J., Babbush, R., Aspuru-Guzik, A.: The theory of variational hybrid quantum-classical algorithms. New Journal of Physics  \textbf{18}(2),  023023 (2016). \doi{10.1088/1367-2630/18/2/023023}

\bibitem{minsky1988}
Minsky, M.L., Papert, S.A.: Perceptrons: An Introduction to Computational Geometry. MIT Press, expanded edition edn. (1988)

\bibitem{nielsen2010}
Nielsen, M.A., Chuang, I.L.: Quantum Computation and Quantum Information. Cambridge University Press, 10th anniversary edition edn. (2010)

\bibitem{preskill2018nisq}
Preskill, J.: Quantum computing in the nisq era and beyond. Quantum  \textbf{2}, ~79 (2018). \doi{10.22331/q-2018-08-06-79}

\bibitem{qiskit}
{Qiskit contributors}: Qiskit. \url{https://qiskit.org/} (2026), accessed: 2026-02-09

\bibitem{schuld2018supervised}
Schuld, M., Petruccione, F.: Supervised Learning with Quantum Computers. Springer, Cham (2018). \doi{10.1007/978-3-319-96424-9}

\bibitem{schuld2014}
Schuld, M., Sinayskiy, I., Petruccione, F.: An introduction to quantum machine learning. Contemporary Physics  \textbf{56} (2014). \doi{10.1080/00107514.2014.964942}

\bibitem{wiki_bloch}
{Wikipedia contributors}: Bloch sphere. \url{https://en.wikipedia.org/wiki/Bloch_sphere} (2026), accessed: 2026-01-03

\end{thebibliography}

\newpage

\appendix
\section{Supplementary Material and Reproducibility}
\label{sec:appendix}

This appendix provides supplementary analyses, visualizations, and
implementation details that complement the main text and support the reported
experimental findings.

All experiments reported in this work are fully reproducible.
The complete source code for dataset generation, model training, evaluation,
and figure reproduction is publicly available in a dedicated GitHub repository.

\medskip
\noindent
\textbf{Repository:} \\
\label{sec:code}
\url{https://github.com/mseilkhan/XOR-research-Quantum-ML-vs-Classic.git}

\medskip
\noindent
The repository contains implementations of all classical and quantum models,
scripts to reproduce the reported figures and tables, and configuration files
specifying datasets, hyperparameters, and random seeds.
All experiments follow fixed data splits and multiple random initializations, as
described in Section~\ref{sec:methodology}.

\subsection{MLP width ablation}
\label{app:mlp_width_ablation}

To motivate the choice of the classical MLP baseline, we conduct a width
ablation over hidden-layer sizes $h \in \{1,2,4,8\}$ on a representative
configuration of Dataset~B ($\sigma=0.10$, $n_{\text{per cluster}}=100$).
All runs follow the same training protocol as in the main text, using a fixed
train/test split (seed~42) and five random initializations.
Results are reported as mean $\pm$ standard deviation.

\begin{figure}[h]
  \centering
  \begin{subfigure}[t]{0.49\textwidth}
    \centering
    \includegraphics[width=\linewidth]{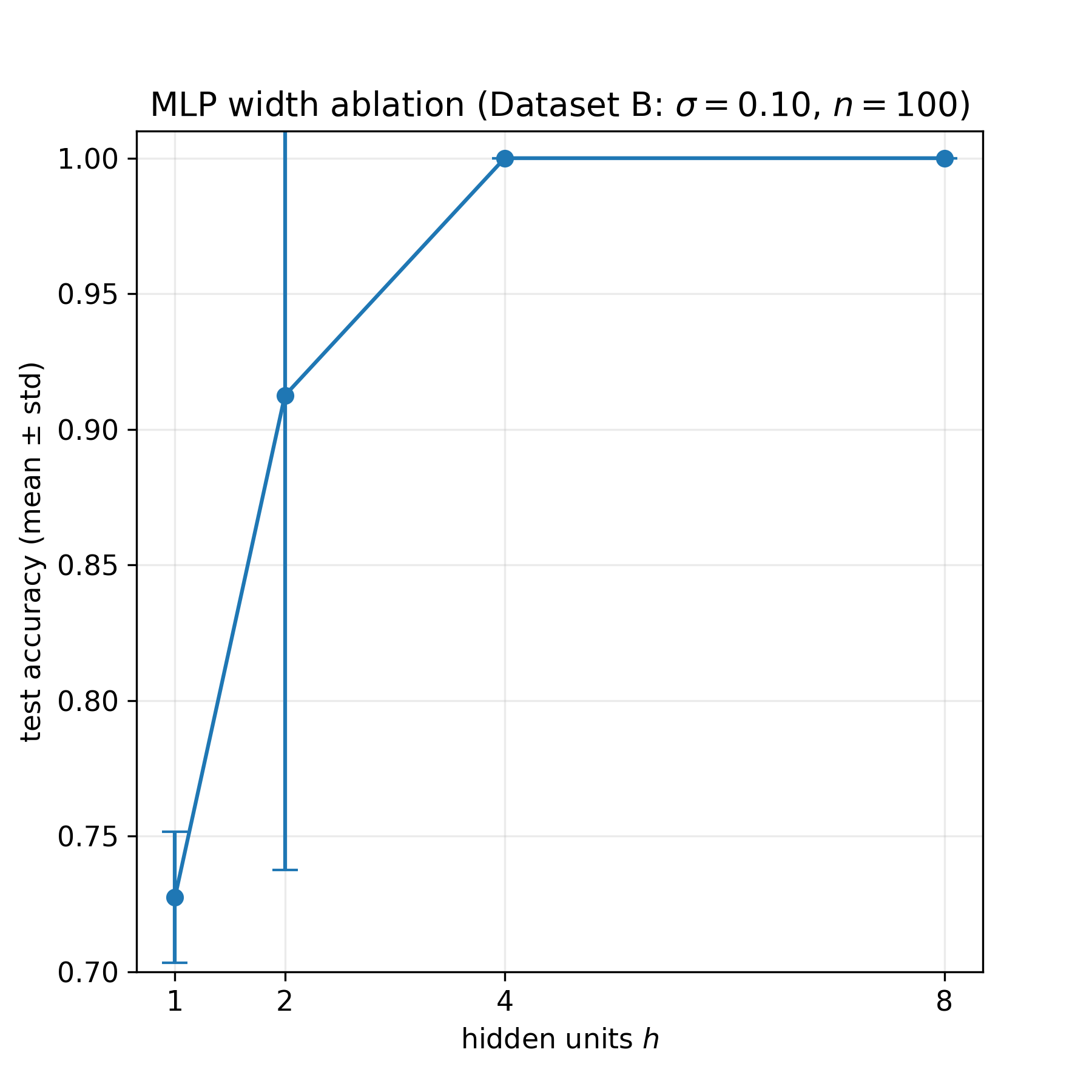}
    \caption{Test accuracy vs.\ width \(h\).}
    \label{fig:mlp_width_acc}
  \end{subfigure}
  \hfill
  \begin{subfigure}[t]{0.49\textwidth}
    \centering
    \includegraphics[width=\linewidth]{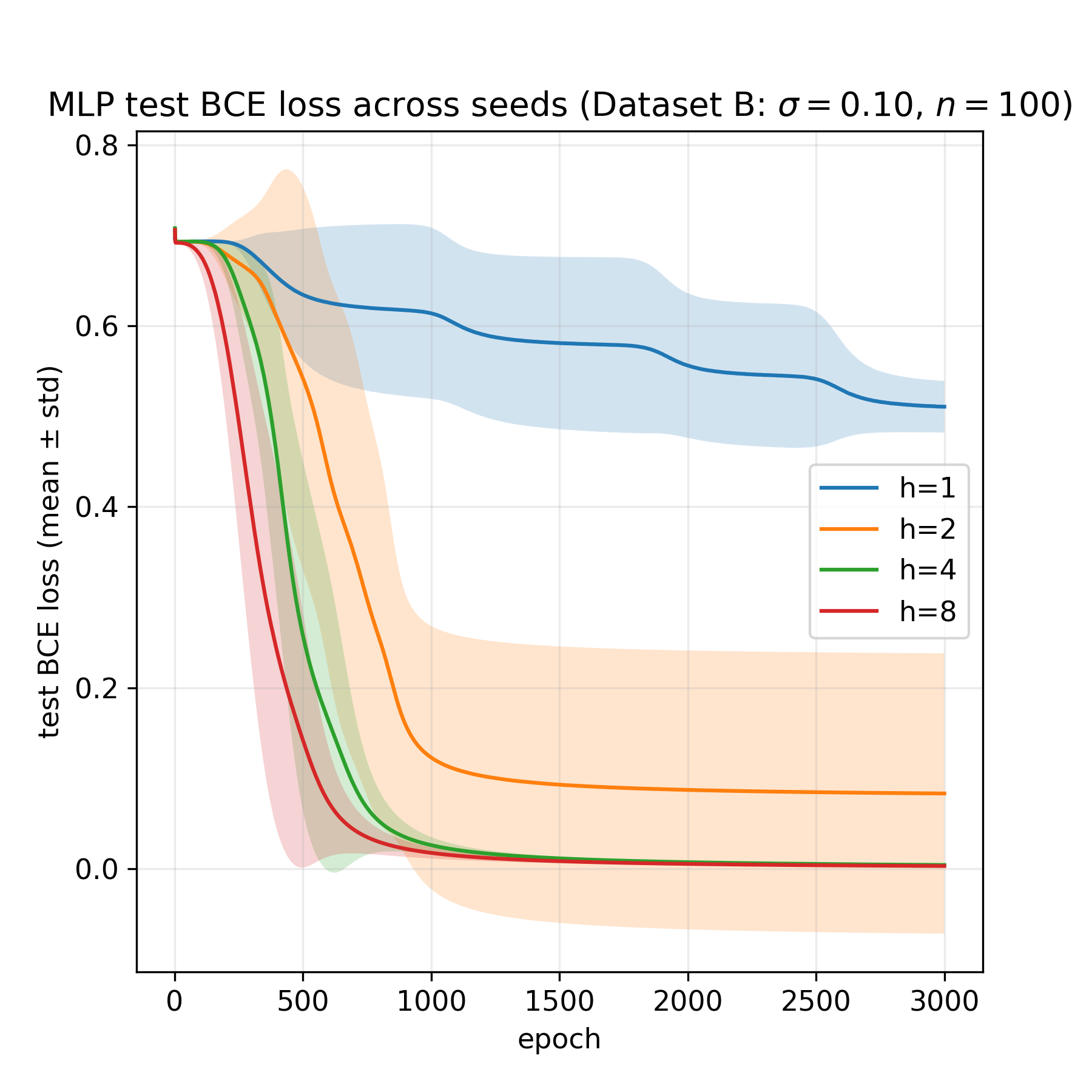}
    \caption{Test BCE loss across epochs.}
    \label{fig:mlp_width_loss}
  \end{subfigure}
  \caption{MLP width ablation on Dataset~B
  (\(\sigma=0.10\), \(n_{\text{per\_cluster}}=100\)).
  Performance improves with increasing width and saturates for moderate model sizes.}
  \label{fig:mlp_width_ablation}
\end{figure}

As shown in Fig.~\ref{fig:mlp_width_ablation}, very small networks (e.g., $h=1$)
fail to reliably capture the XOR structure under noise.
Increasing the hidden width improves accuracy and training stability, with
performance saturating for $h \ge 4$.
Accordingly, we fix the MLP baseline to $h=4$ in all main experiments, as it
provides sufficient expressivity while remaining compact and comparable to the
quantum classifiers.

\subsection{Loss Landscape Slices}
\label{app:loss_landscape}

\begin{figure}[h]
\centering

\begin{subfigure}[t]{0.49\textwidth}
  \centering
  \includegraphics[width=\linewidth]{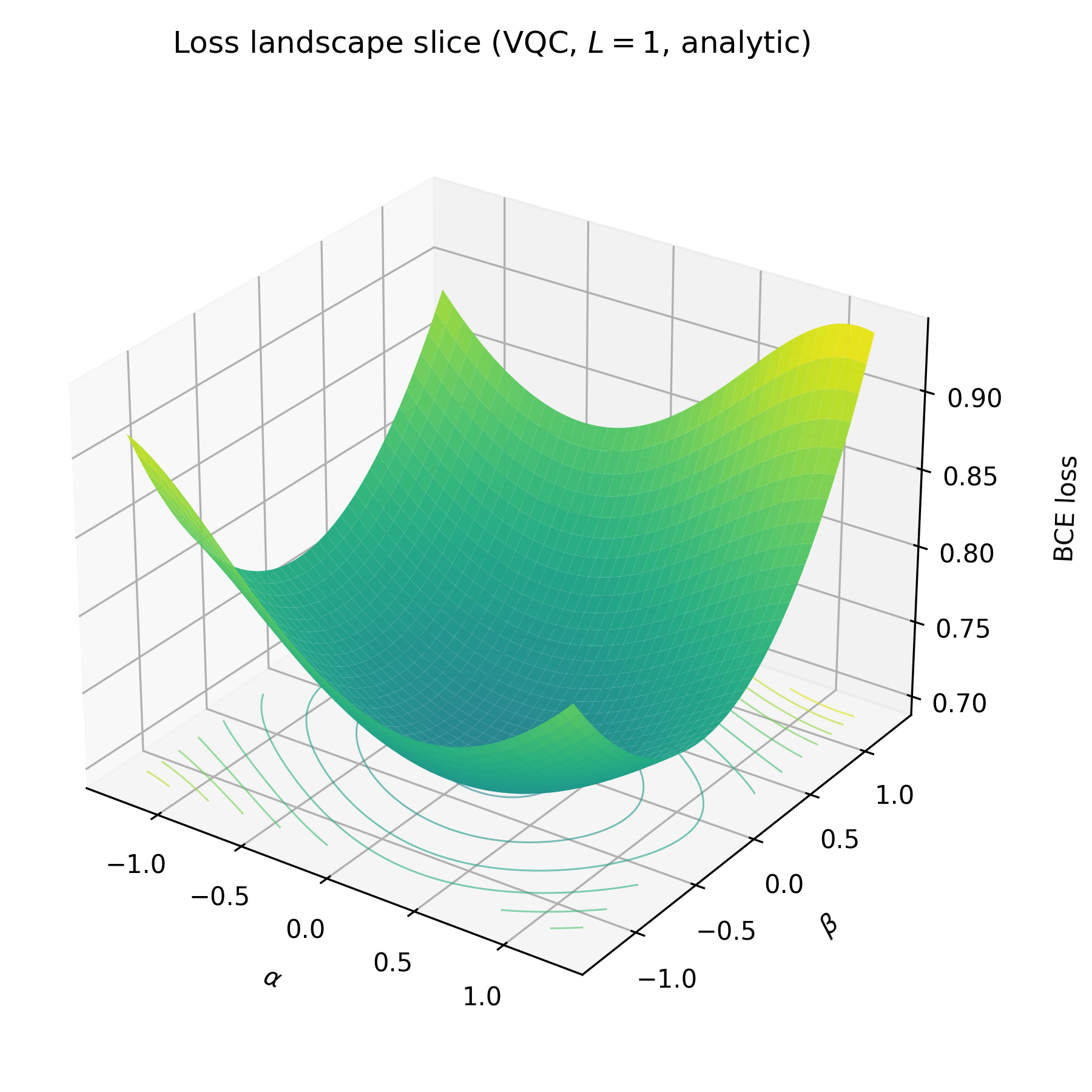}
  \caption{VQC ($L=1$, analytic)}
  \label{fig:app_loss_landscape_L1}
\end{subfigure}
\hfill
\begin{subfigure}[t]{0.49\textwidth}
  \centering
  \includegraphics[width=\linewidth]{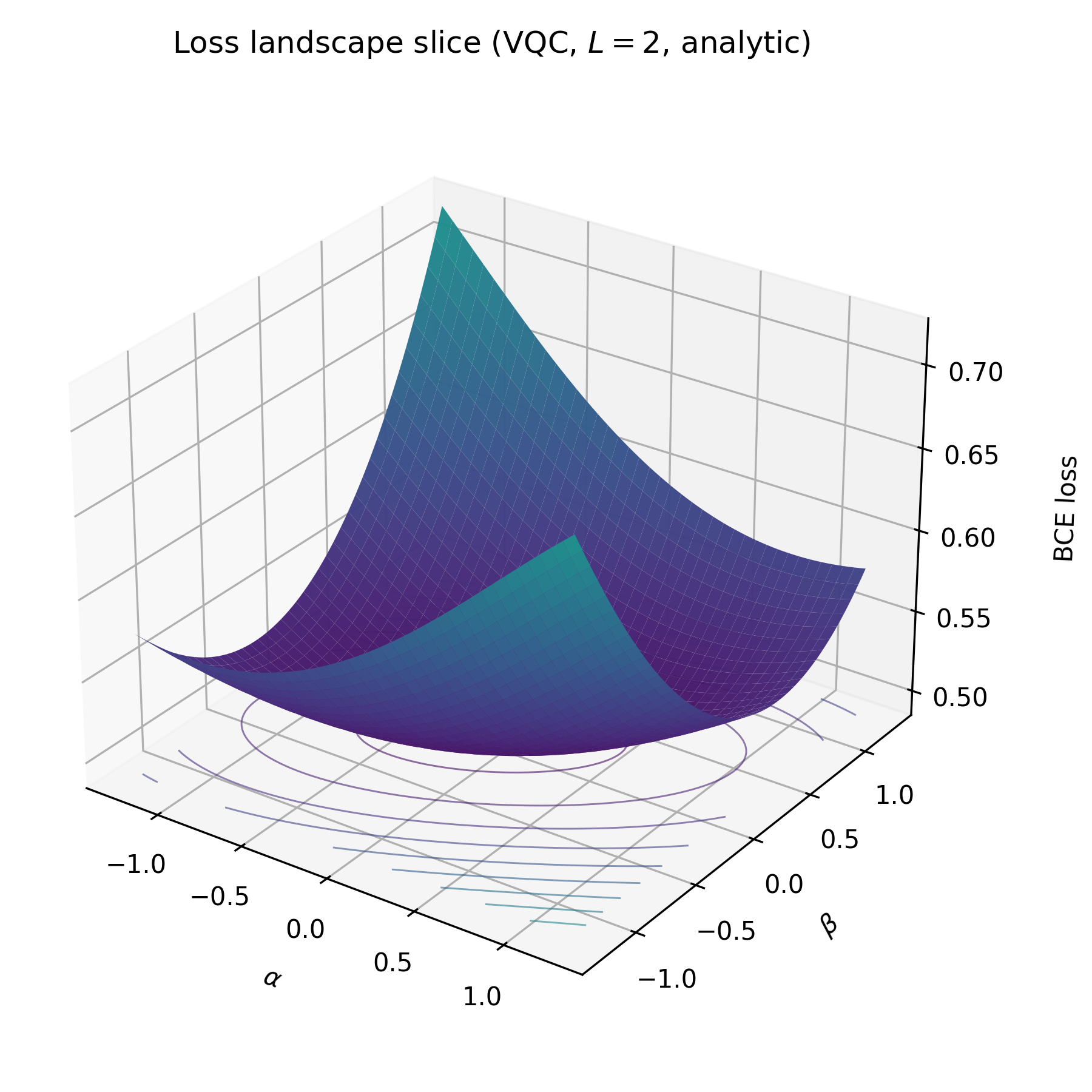}
  \caption{VQC ($L=2$, analytic)}
  \label{fig:app_loss_landscape_L2}
\end{subfigure}

\vspace{0.6em}

\includegraphics[width=0.68\linewidth]{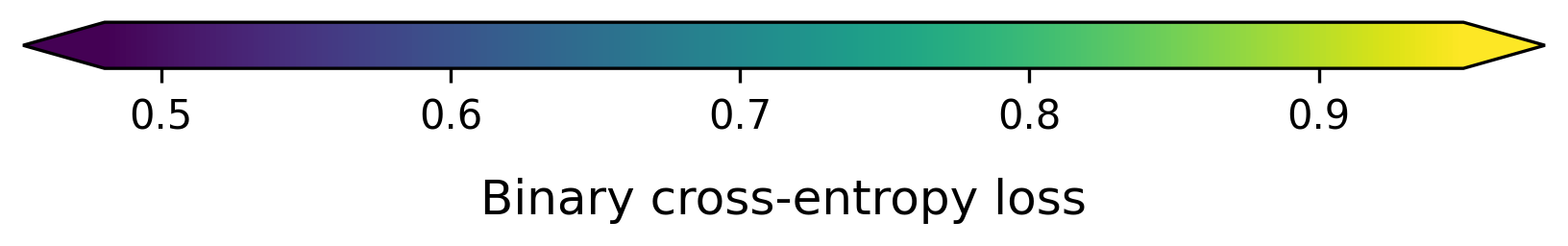}

\caption{
Two-dimensional parameter-space loss landscape slices in parameter space for the variational quantum classifier at circuit depths $L=1$ and $L=2$ (analytic regime). 
The horizontal color scale indicates the binary cross-entropy loss value and is shared across both surfaces.
}
\label{fig:app_loss_landscape}
\end{figure}

Each surface represents a two-dimensional slice of the binary cross-entropy loss
as a function of two selected circuit parameters, with all remaining parameters
fixed at their trained values.
Lower regions correspond to better-performing parameter configurations, while
the curvature and spread of the surface provide qualitative information about
local smoothness and optimization sensitivity.

\newpage

\subsection{Additional Hardware Diagnostics}
\label{app:hardware_additional}

\begin{figure}[h]
\centering
\includegraphics[width=0.60\linewidth]{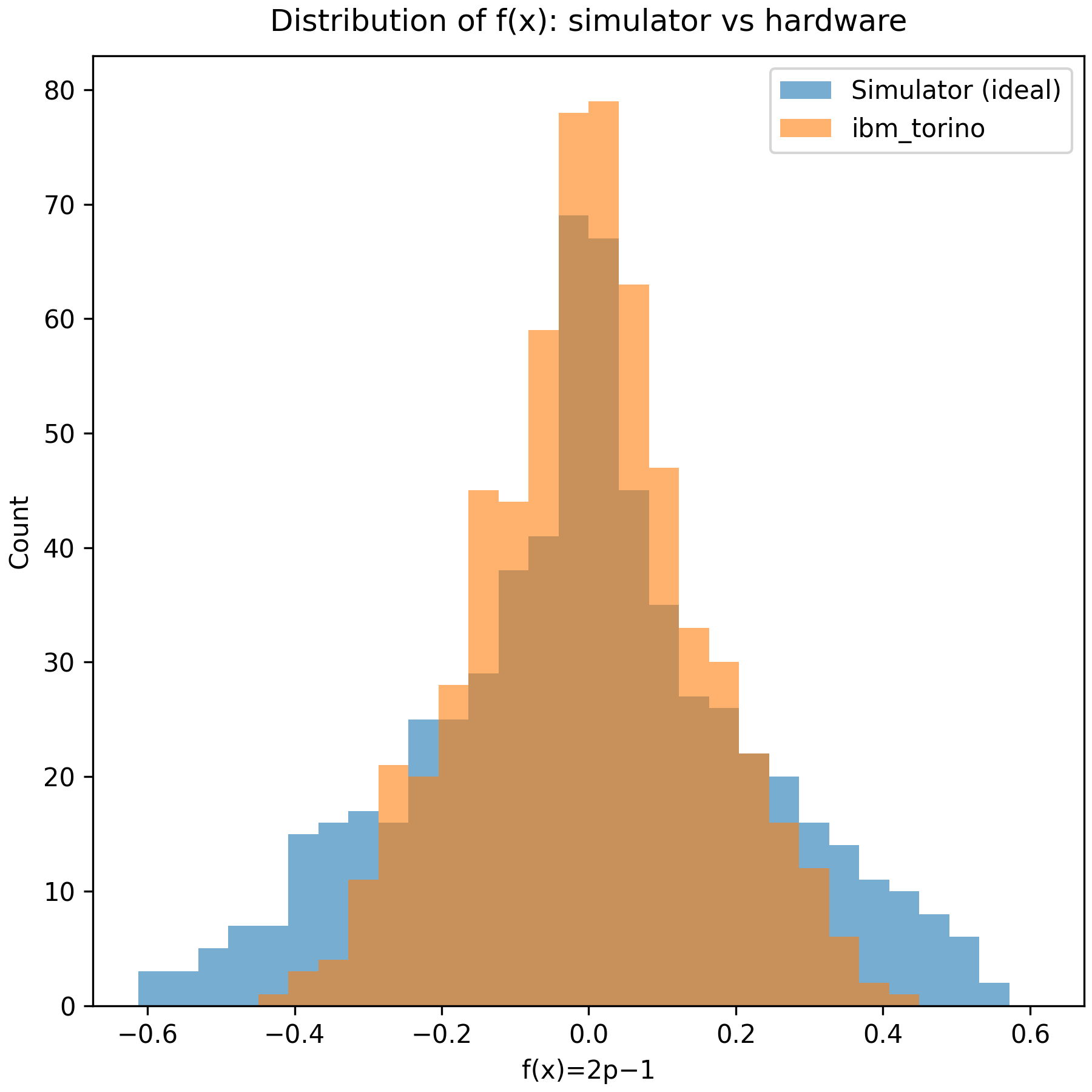}
\caption{Distribution of decision scores $f(x)=2p-1$ over the evaluation grid:
ideal simulation vs.\ IBM hardware. Hardware execution shows a visible compression
toward zero, consistent with reduced contrast caused by device noise.}
\label{fig:hw_hist}
\end{figure}

\subsection{Dataset C: Continuous XOR}
\label{app:datasetC}

This appendix reports qualitative results for Dataset~C, a continuous
threshold-based variant of the XOR problem defined in
Section~\ref{sec:methodology}, with emphasis on learning dynamics and the
dependence of performance on the threshold parameter~$t$.

\paragraph{Learning dynamics.}
Figure~\ref{fig:appC_lc} shows training behavior at $t=0.5$ using median test BCE
over random initializations, with the interquartile range (IQR) indicating
variability.
The variational quantum classifier with two layers, VQC($L{=}2$), rapidly reaches
a stable loss regime and exhibits consistently low dispersion across seeds.
In contrast, the classical MLP shows increased variability at early training
stages, attributable to optimization dynamics under a fixed training budget and
not necessarily indicative of its asymptotic behavior.

All results correspond to predefined training horizons and therefore reflect
optimization behavior under constrained training budgets rather than asymptotic
convergence.

\paragraph{Dependence on threshold $t$.}
Figure~\ref{fig:appC_threshold}(b, c) summarizes test accuracy and test BCE as functions
of the threshold~$t$.
All models perform worst near $t \approx 0.5$, where the decision boundary is
maximally symmetric.
Across the full sweep, VQC($L{=}2$) achieves the highest accuracy and lowest BCE
with comparatively small variability, reflecting the increased expressive
capacity of the deeper variational ansatz on smooth XOR-like decision surfaces.

\begin{figure}[!htbp]
  \centering
  \begin{subfigure}[t]{0.46\linewidth}
    \centering
    \includegraphics[width=0.95\linewidth]{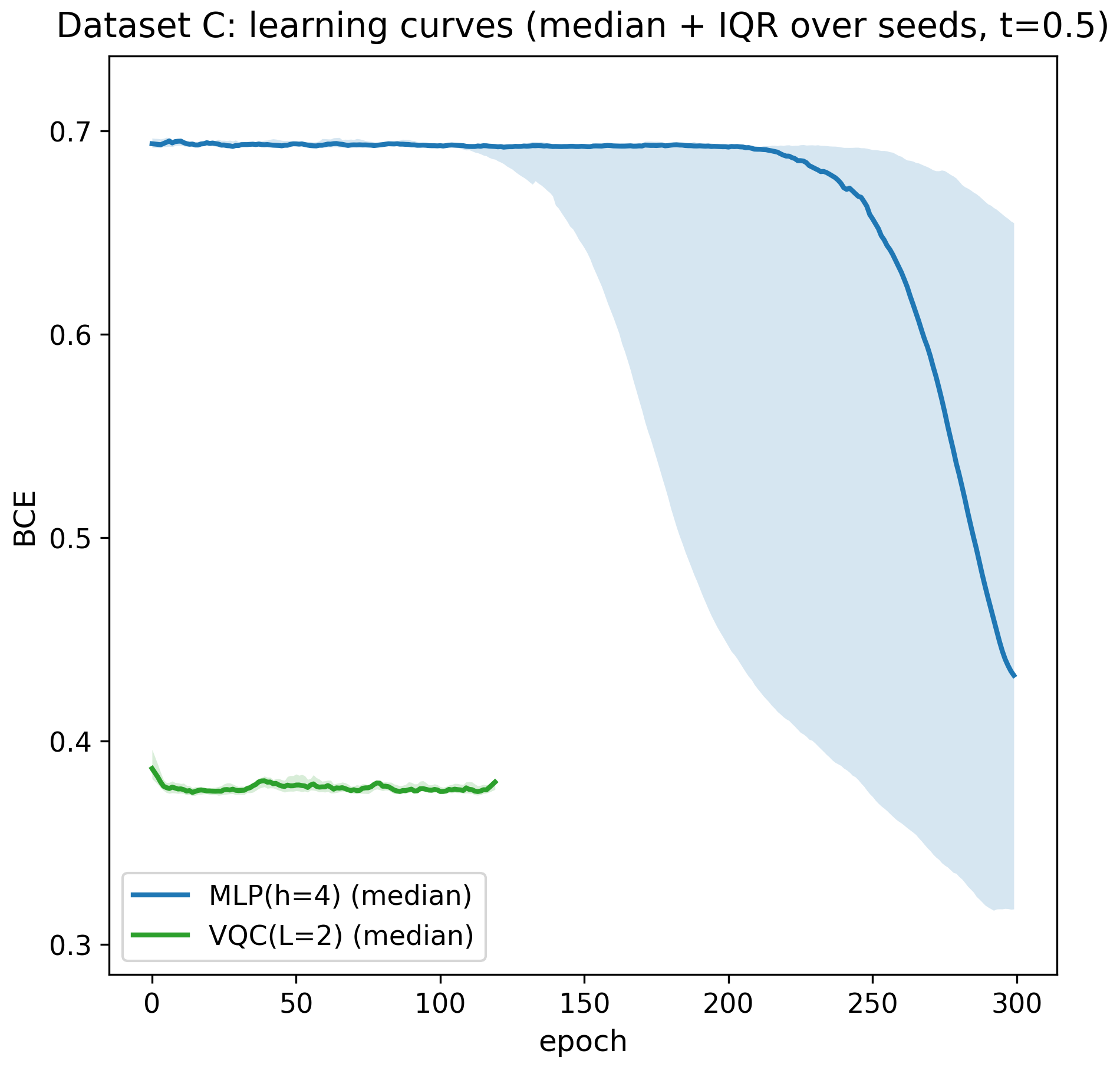}
    \caption{Learning curves at $t=0.5$ (median test BCE with IQR).}
    \label{fig:appC_lc}
  \end{subfigure}
  \hfill
  \begin{subfigure}[t]{0.46\linewidth}
    \centering
    \includegraphics[width=0.95\linewidth]{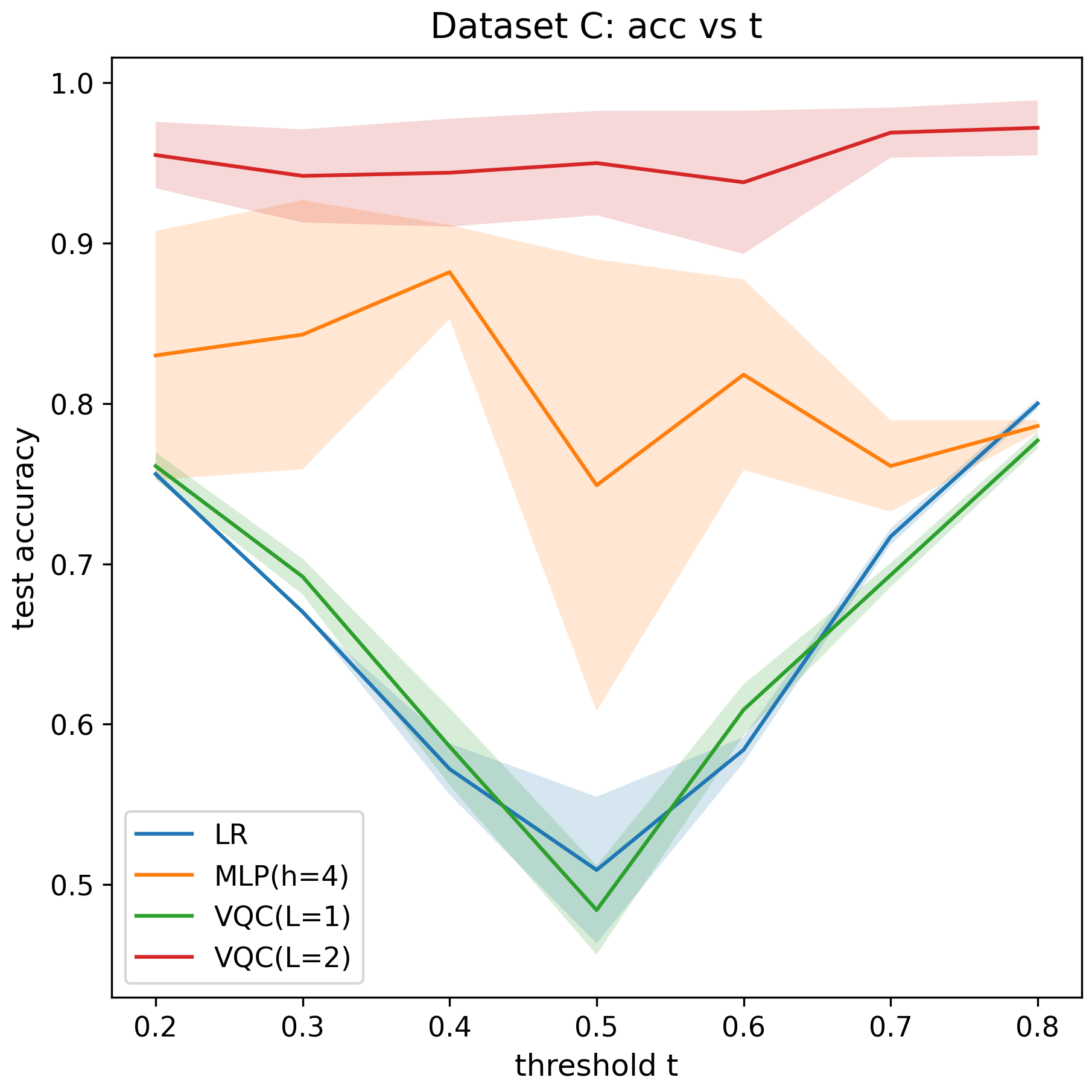}
    \caption{Test accuracy as a function of threshold $t$.}
    \label{fig:appC_acc_t}
  \end{subfigure}

  \vspace{0.3em}

  \begin{subfigure}[t]{0.46\linewidth}
    \centering
    \includegraphics[width=0.95\linewidth]{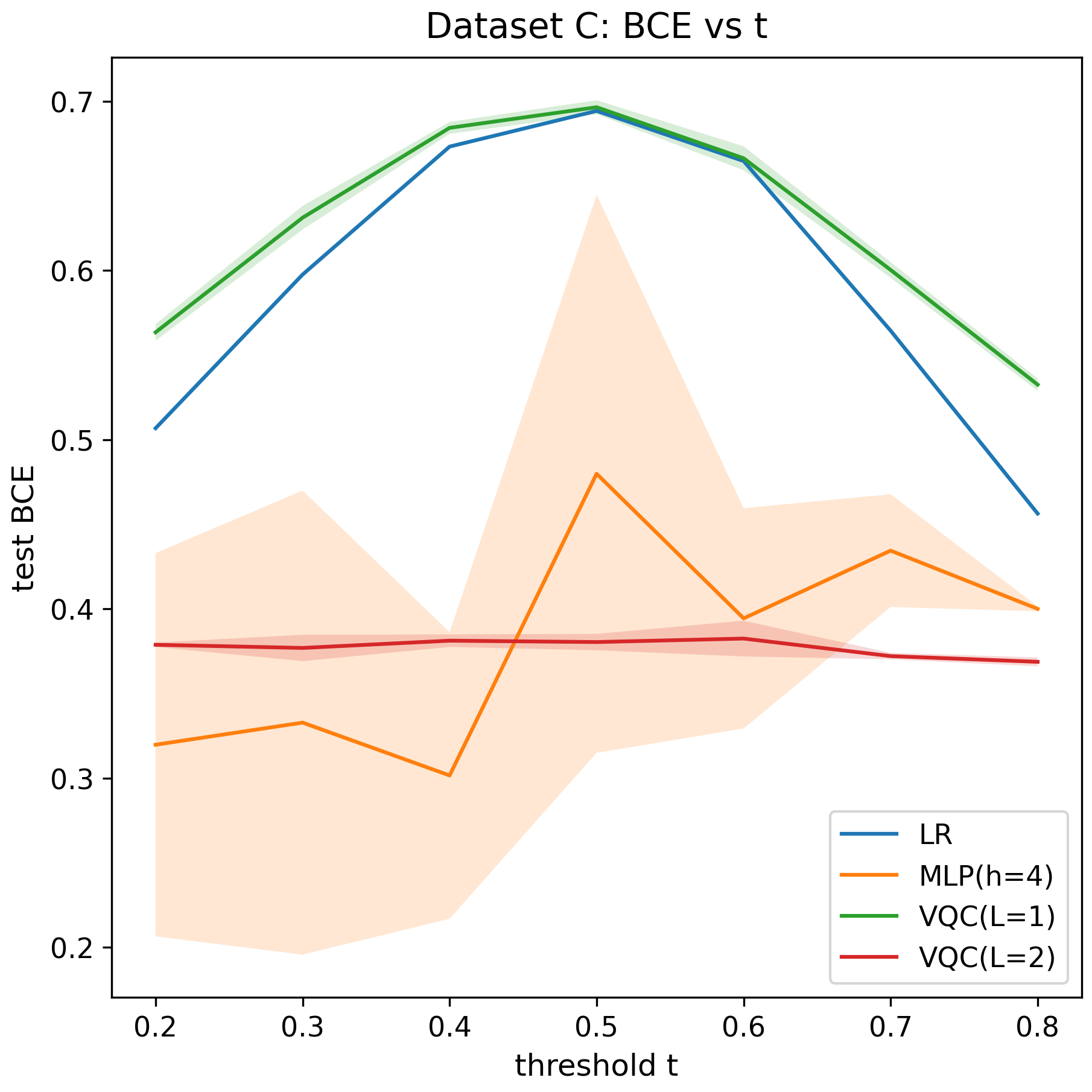}
    \caption{Test BCE as a function of threshold $t$.}
    \label{fig:appC_bce_t}
  \end{subfigure}

  \caption{Qualitative results for Dataset~C: learning dynamics and threshold
  dependence across classical and variational models.}
  \label{fig:appC_threshold}
\end{figure}

\end{document}